\pdfoutput=1
\documentclass[11pt]{article}

\usepackage[]{acl}

\usepackage{times}
\usepackage{latexsym}
\usepackage{multirow}
\usepackage{graphicx}
\usepackage{booktabs}
\usepackage{tcolorbox}
\usepackage{multirow}
\usepackage{lscape}
\usepackage{arydshln}
\usepackage{microtype}
\usepackage[normalem]{ulem}
\useunder{\uline}{\ul}{}
\newcommand{\myexample}[2]{
    \begin{tcolorbox}[colback=white!5!white,colframe=black,title={Example: #1}]
        #2
    \end{tcolorbox}
}

\usepackage[T1]{fontenc}
\usepackage{authblk}

\usepackage[utf8]{inputenc}

\usepackage{microtype}

\usepackage{inconsolata}

\title{Assessing Translation capabilities of Large Language Models involving English and Indian Languages}

\author[ ]{Vandan Mujadia}
\author[ ]{Ashok Urlana}
\author[ ]{Yash Bhaskar}
\author[ ]{\\Penumalla Aditya Pavani}
\author[ ]{Kukkapalli Shravya}
\author[ ]{\\Parameswari Krishnamurthy}
\author[ ]{Dipti Misra Sharma}
\affil[ ]{LTRC, IIIT Hyderabad, India}
\affil[ ]{\{vandan.mu,ashok.urlana,yash.bhaskar\}@research.iiit.ac.in,}
\affil[ ]{\{aditya.pavani,kukkapalli.shravya\}@students.iiit.ac.in, \{param.krishna,dipti\}@iiit.ac.in}

\begin{document}
\maketitle
\begin{abstract}
%Generative Large Language Models (LLMs) have achieved remarkable advancements in various NLP tasks. In this work, we aim to explore the multilingual capabilities of large language models using machine translation as a task involving English and 22 Indian languages. We first explore the translation capabilities of raw large language models, followed by in-context learning capabilities of the same raw models, and finally, we fine-tune these large language models using parameter efficient fine-tuning methods such as LoRA and full fine-tuning. Through our study, we have identified the best performing large language model for the translation task involving LLMs, which is based on LLaMA. 
%Our results demonstrate significant progress, with average BLEU scores of 28.47, 28.87, 26.29, 20.13 and 13.92 as well as CHRF scores of 43.02, 47.76, 38.94, 37.04 and 31.47 for English to Indian languages on IN22 (conversational), IN22 (general), flores200-dev, flores200-devtest, newstest2019 testsets respectively. Similarly, for Indian languages to English, we achieved average BLEU scores of 25.94, 18.16, 38.21, 24.24 and 10.26 along with CHRF scores of 36.51, 32.98, 53.77, 33.74 and 18.19 on IN22 (conversational), IN22 (general), flores200-dev, flores200-devtest, newstest2019 testsets respectively. Overall, our findings highlight the potential and strength of large language models for machine translation capabilities for languages that are currently underrepresented in LLMs. 

Generative Large Language Models (LLMs) have achieved remarkable advancements in various NLP tasks. In this work, our aim is to explore the multilingual capabilities of large language models by using machine translation as a task involving English and 22 Indian languages. We first investigate the translation capabilities of raw large language models, followed by exploring the in-context learning capabilities of the same raw models. We fine-tune these large language models using parameter efficient fine-tuning methods such as LoRA and additionally with full fine-tuning. Through our study, we have identified the best performing large language model for the translation task involving LLMs, which is based on LLaMA.

Our results demonstrate significant progress, with average BLEU scores of 13.42, 15.93, 12.13, 12.30, and 12.07, as well as CHRF scores of 43.98, 46.99, 42.55, 42.42, and 45.39, respectively, using 2-stage fine-tuned LLaMA-13b for English to Indian languages on IN22 (conversational), IN22 (general), flores200-dev, flores200-devtest, and newstest2019 testsets. Similarly, for Indian languages to English, we achieved average BLEU scores of 14.03, 16.65, 16.17, 15.35 and 12.55 along with chrF scores of 36.71, 40.44, 40.26, 39.51, and 36.20, respectively, using fine-tuned LLaMA-13b on IN22 (conversational), IN22 (general), flores200-dev, flores200-devtest, and newstest2019 testsets. Overall, our findings highlight the potential and strength of large language models for machine translation capabilities, including for languages that are currently underrepresented in LLMs.

\end{abstract}

\section{Introduction}

Generative Large Language Models (LLMs) have made significant performance improvements in various natural language processing (NLP) tasks, showcasing exceptional progress in a wide range of applications \citep{xuanfan-piji-2023-systematic,xi2023rise}. These tasks span from open domain question answering, where LLMs excel at providing accurate and coherent responses, to instruction-based tasks such as code completion, where LLMs can generate code snippets based on given prompts \citep{vaithilingam2022expectation}. LLMs have also demonstrated proficiency in tasks like essay writing, grammar checking \citep{wu2023chatgpt}, and text summarization, where they can produce high-quality outputs \citep{chang2023survey}. These advancements have primarily been observed in English-centric tasks. The popular LLMs support several of natural languages. The performance for some languages other than English is not yet on par or yet to be evaluated \citep{lai2023chatgpt,zhu2023multilingual}.\\

\begin{figure}
    \centering
    \includegraphics[width=220pt]{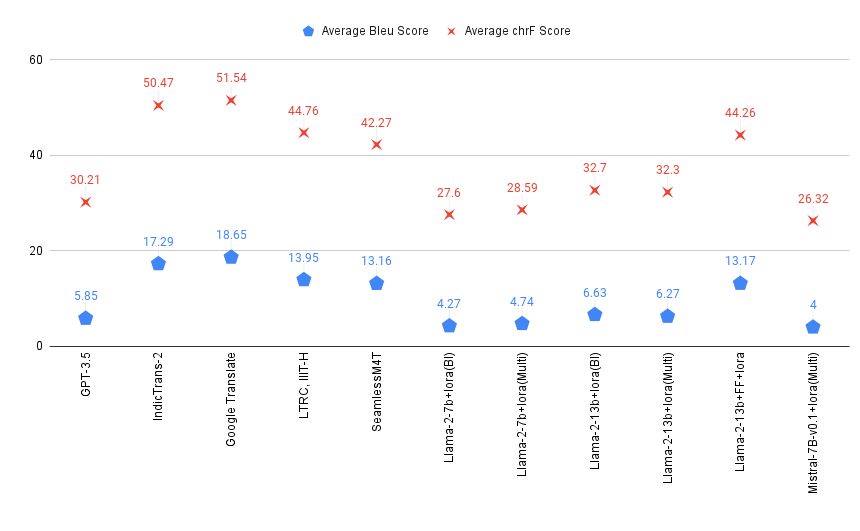}
    \caption{LLMs based Machine Translation performance comparison with public systems for \textbf{English to Indian Languages}. BLEU and chrF scores are averaged over 22 Indian Languages and 5 different benchmark data-sets. The available MT systems are GPT-3.5 (GPT-3.5 Davinci, by OpenAI), IndicTrans-2, Google Translation, LTRC-IIIT-H, SeamlessM4T. LLaMA-2-7b and LLaMA-2-13b are evaluated as LLM based fine-tuned MT systems are namely LLaMA-2-7b+lora (Multi), LLaMA-2-13b+lora (Multi), and LLaMA-2-13b+FF+lora (Multi).}
    \label{fig:mainresults1}
\end{figure}
A multilingual country like India, where over 364+ languages and dialects \footnote{\url{https://en.wikipedia.org/wiki/Linguistic_Survey_of_India}} are spoken across its vast territory, presents a multitude of challenges across various domains due to language barriers \citep{zielinski2021language}, such as day-to-day communication, education \citep{steigerwald2022overcoming}, business, healthcare \citep{mehandru2022reliable}, tourism, governance, and more. Recent advancements in the field of Large Language Models may offer solutions to these challenges tailored to Indian languages. \\

%To effectively address language barriers with large language models, it is crucial to assess its proficiency in multilinguality (handling Indian languages) as a first step. Machine Translation, as one of the crucial multilingual tasks, has the potential to bridge language gaps, facilitate effective communication, and enhance accessibility to resources in multiple languages \citep{lin1999machine}. This makes it a perfect task to explore the multilingual capabilities of existing large language models. 

%To effectively address language barriers with large language models, it is crucial to assess their proficiency in handling Indian languages. Machine Translation, as a crucial multilingual task, can be an ideal task to explore the multilingual capabilities of existing large language models. Therefore, we can frame and ask the question as: \textbf{How well do Large Language Models perform on a multi-lingual task such as Machine Translation involving Indian Languages?}\\

To test whether LLM can effectively overcome language barriers, it is crucial to evaluate the proficiency of large language models in handling Indian languages. Machine Translation, as a critical multilingual task, could be an ideal option to explore the multilingual capabilities of existing models. Hence, we can formulate the question to assess the proficiency of large language models in handling Indian languages as follows: \textbf{How effectively do large language models perform in multilingual tasks like Machine Translation, particularly when dealing with Indian languages?}\\

In this work, our major contribution is to address the following points in response to the above question.
\begin{itemize}
    \item What are the directions for utilizing or adapting Large Language Models for Indian Languages?
    \begin{itemize}
        \item How do LLMs perform in translating a wide range of Indian languages under zero-shot and in-context learning settings?
        \item Does LLM fine-tuning improve the translation capabilities of Large Language Models? How do they perform on low-resourced MT languages?
        \item The impact of LLM Vocabulary on the Performance of Large Language Models in Translation Tasks.
    \end{itemize}
\end{itemize}
To address the above questions, we assess the translation capabilities of popular large language models (opt, bloom, LLaMA-1, MPT, Falcon, LLaMA-2, and Mistral [Section \ref{LLMb}]) involving English and 22 scheduled Indian languages (Assamese, Bangla, Bodo, Dogri, Konkani, Gujarati, Hindi, Kannada, Kashmiri, Maithili, Malayalam, Marathi, Meitei, Nepali, Odia, Punjabi, Sanskrit, Santali, Sindhi, Tamil, Telugu, and Urdu\label{langall}). We initially examine the translation capabilities of above mentioned raw large language models [Section \ref{LLMPromp1}]. Subsequently, we explore their in-context learning abilities [Section \ref{LLMPromp2}]. Additionally, we fine-tune the base models using parameter-efficient fine-tuning methods specifically LoRa [Section-\ref{LLMPromp3}]. Furthermore, we investigate the potential of 2-stage fine-tuning for large language models, which involves full parameter fine-tuning in the first stage, followed by LoRa-based adaptor fine-tuning [Section \ref{LLMPromp4}].\\

The key findings of our work, as summarized in Figure~\ref{fig:mainresults1}, highlight the performance of our LLM-based machine translation fine-tuned models compared to various known translation engines. These engines range from commercial (Google\footnote{\url{https://translate.google.co.in/}}, GPT-3.5\footnote{\url{https://chat.openai.com/}}) to open source (IndicTrans-2\footnote{\url{https://github.com/AI4Bharat/IndicTrans2}}, LTRC-IIIT-H\footnote{\url{https://ssmt.iiit.ac.in/translate}}, seamlessm4t\footnote{\url{https://github.com/facebookresearch/seamless\_communication}}), traditional supervised encoder-decoder translation models (Google, IndicTrans-2, LTRC-IIIT-H) and decoder-driven causal large language model-based translation systems (GPT-3.5). \\

Our findings underscore the significant potential of large language models for translation tasks involving English and Indian Languages. While raw LLMs (LLaMA-2-7b and LLaMA-2-13b) not perform well on translation tasks, our two-stage MT fine-tuned models (LLaMA-2-13b+FF+lora(Multi)) yields comparative results even with minimal parallel corpora. This suggests that LLMs have the potential to possess multilingual capabilities for translating into underrepresented languages, which can be further enhanced through fine-tuning. This work will be a crucial and pioneering milestone in evaluating LLMs for language representation and assessing their translation capabilities for a diverse range of Indian languages, especially those with limited available resources.

\section{Related Work}
Recent advancements in machine translation have shown that neural machine translation (NMT) has made significant strides in terms of output fluency and translation quality, especially when ample parallel data is available \citep{wmt-2020-machine}. However, the scarcity or absence of parallel data poses a challenge for most language pairs. In the case of Indian languages, recent developments have tried to addressed this issue by introducing a new state-of-the-art approach: multilingual machine translation involving Indian languages and English \citep{wang2021survey,dabre2020survey}. This approach leverages a single script for machine translation, capitalizing on the lexical and syntactic similarities that arise from the genetic and contact-relatedness among Indian languages \citep{gala2023indictrans2,eriguchi2022building,bapna2019exploring}.\\

In the field of LLM driven machine translation, in-context learning has gained significant attention \citep{wu-etal-2023-openicl}. The use of large language models (LLMs) for multilingual machine translation has been a subject of interest \citep{zhang2023prompting}. Recent studies have evaluated the translation capabilities of LLMs for different language directions, with a focus on models like ChatGPT \citep{bang2023multitask}. Notably, \citeauthor{xu2023paradigm} proposed a two-stage fine-tuning approach for machine translation using LLMs, involving fine-tuning on monolingual data followed by fine-tuning on a small set of high-quality parallel data. Our work represents the first study that specifically explores machine translation involving Indian languages using large language models.

\section{Large Language Models}
\label{LLMb}
Language modeling, a well-established task in the field of natural language processing, has garnered significant attention over the years \citep{bellegarda2004statistical, bengio2000neural}. This task involves predicting the probability of the next token in a sequence of words. Transformers have emerged as the fundamental architecture underlying many existing Large Language Models \citep{vaswani2017attention}. \\

Transformers based autoregressive models like GPT \cite{brown2020language, radford2019language} have played a crucial role in advancing Natural Language Processing (NLP). GPT-3, with 175 billion parameters, is a standout in this category. It is similar in structure to GPT-2 and GPT-1 but benefits from a more extensive and varied dataset, making it exceptionally powerful in NLP. Further, prompt-based ChatGPT (GPT-3.5 text-davinci-003 and GPT-3.5 turbo) has been performing exceptionally by utilizing the reinforcement-based human feedback strategy. Although these models exhibit impressive performance on several NLP tasks, privacy and bias of the models have been a bottleneck. To mitigate such issues, LLaMA \cite{touvron2023LLaMA} is an open-sourced foundation model trained on publicly available datasets. Similarly, Falcon-40B \cite{almazrouei2023falcon} is another open-source LLM trained on a RefinedWeb corpus of 1500 billion tokens. Falcon even comes with 7 and 40 billion instruction versions trained on conversation data.\\

%Recent adaptation of LLMs for instruction tuning has been a way to advance the performance of many natural language processing tasks. Recent adaption of instruction tuning of Chinese \cite{cui2023efficient} and Swedish \cite{holmstrom-doostmohammadi-2023-making} proves the zero-shot and generation ability of the low-rank adaptation of LLaMA for non-English languages. As of now, these instruction models mainly concentrate on English, there is an immediate requirement to find a way to adapt these models to low-resource Indian languages.

The recent adaptation of Large Language Models (LLMs) for instruction tuning has proven to be a promising approach in improving the performance of various natural language processing tasks. Specifically, in languages like Chinese and Swedish demonstrates the impressive zero-shot and generation abilities of the low-rank adaptation of LLaMA for non-English languages \citep{cui2023efficient,holmstrom-doostmohammadi-2023-making}. However, it is worth noting that the current focus of these instruction models is primarily on English. Therefore, there is an immediate need to explore ways to adapt these models to low-resource Indian languages.

%We used the Stanford alpaca-lora \cite{taori2023stanford} model to fine-tune Indian languages. The LLaMA model was created on the assumption that the best performance was obtained by training a small model with more data. It presents the series of models available from 7B to 65B parameters with competitive performance with state-of-the-art models. LLaMA is pre-trained on 1.4T tokens. The architecture of LLaMA is based on transformers \cite{vaswani2017attention} with several modifications. The modifications include pre-normalization \cite{zhang2019root} of each transformer sub-layer, utilizing the SwiGLU \cite{shazeer2020glu} activation function, rotary embeddings \cite{su2021roformer}, and efficient optimization techniques to improve the training speed and memory usage. Recently, LLaMA2 \cite{touvron2023LLaMA2} extended the max context length of 2046 to 4096 tokens. Unlike the LLaMA model, LLaMA2 uses group-query attention instead of multi-query attention and is trained on 40\% more data than the previous version.

%OPT \cite{zhang2022opt} is another open-source decoder-only model, which shows comparable performance with GPT-3. MPT-7B \cite{team2023introducing} was trained on 1T token of text and code. It matches the quality of LLaMA-7B.   

%\input{llms_lang_support}
\subsection{Base models}
In this work, we used the following base LLM models to test the levels of language coverage and explore their potential for machine translation tasks involving English and Indian languages.
\begin{itemize}
    \item \textbf{opt-6.7b\footnote{\url{https://huggingface.co/facebook/opt-6.7b}}} : The OPT-6.7b \citep{zhang2022opt} model has been extensively trained on the objective of causal language modeling (CLM) using English text. Although the majority of the training data is in English, a small portion of non-English data from CommonCrawl has also been included. This model utilizes 6.7 billion parameters, consisting of 32 layers and 32 attention heads, and employs an embedding size of 4096.
    \item \textbf{Bloom-7B\footnote{\url{https://huggingface.co/bigscience/bloom-7b1}}} : BLOOM \cite{scao2022bloom} was the first largest multilingual large language model with causal language modeling objective and supports 46 languages and 13 programming languages. Its overall training data contains 1.1\% of Indian languages. We opted for Bloom model with 7,069,016,064 parameters with 30 layers, 32 attention heads, 4096 embedding dimensional where maximum token length is 2048.
    \item \textbf{LLaMA-7B\footnote{\url{https://huggingface.co/decapoda-research/llama-7b-hf}}}: LLaMA is a collection of foundation language models ranging from 7B to 65B parameters. These models are multilingual models and trained on trillions of tokens. The data includes CCNet, C4, GitHub, Wikipedia, Books, ArXiv, Stack Exchange. In our experiments we evaluated LLaMA model with 7B parameters where	4096 is embedding dimensions and 32	layers and 32 attention head.
    \item \textbf{MPT-7B\footnote{\url{https://huggingface.co/mosaicml/mpt-7b}}} : Similar to above models MPT-7B model is trained on a large amount of data 1T tokens on causal language modeling objective.
    \item \textbf{Falcon\footnote{\url{https://huggingface.co/tiiuae/falcon-7b}}} : Falcon \citep{refinedweb} is another large language model trained on causal language modeling (CLM) objective. Here, we utilised Falcon-7B model which is a 7B parameters and trained on 1.5 trillion tokens of RefinedWeb (a novel massive web data-set based on CommonCrawl) enhanced with curated corpora. The model has multilingual capabilities but no Indian languages are explicitly present. We have used Falcon-7B for our experiments. 
    \item \textbf{LLaMA-2-7B\footnote{\url{https://huggingface.co/meta-llama/Llama-2-7b-hf}} and LLaMA-2-13B\footnote{\url{https://huggingface.co/meta-llama/Llama-2-13b-hf}}} : LLaMA 2 based models \cite{touvron2023LLaMA2} are also trained on causal language modeling (CLM) objective and pretrained on 2 trillion tokens of data from publicly available sources of till September 2022. These models are available in different range parameters from 7 billion to 70 billion. These models have 4k sub-words as context length. In our experiments we have experimented with 7B and 13B LLaMA-2 models. LLaMA-2-7B network has 32 layers and 32 attention heads while LLaMA-2-13B has 40 layers and 40 attention heads.
    \item \textbf{Mistral-7B\footnote{\url{https://huggingface.co/mistralai/Mistral-7B-v0.1}}} : Mistral-7B Large Language Model (LLM) \citep{jiang2023mistral} is a pre-trained on causal language modeling (CLM) objective with 7 billion parameters. It uses Sliding Window Attention (SWA) to handle longer sequences at smaller cost and Grouped-query attention (GQA) for faster inference which reduces the memory requirement during decoding. It has 4096 embedding dimension, 32 layers and 32 attention heads with context length of 8192 context length.  
\end{itemize}

\section{Indian Languages representation in LLMs}
% Please add the following required packages to your document preamble:
% \usepackage{graphicx}
% \usepackage[normalem]{ulem}
% \useunder{\uline}{\ul}{}
\begin{table*}[h]
\centering
\resizebox{2.05\columnwidth}{!}{%
\begin{tabular}{lcccccccccccccll|ccll|cc|c}
\toprule
\textbf{Language Family} & \multicolumn{15}{c|}{{\ul \textbf{Indo-Aryan}}}                                                                                                                                                                                                                                  & \multicolumn{4}{c|}{{\ul \textbf{Dravidian}}}                                                               & \multicolumn{2}{c|}{{\ul \textbf{Sino-Tibetan}}} & {\ul \textbf{Austroasiatic}} \\
\textbf{Language}        & \textbf{asm} & \textbf{ban} & \textbf{kas} & \textbf{snd} & \textbf{urd} & \textbf{doi} & \textbf{hin} & \textbf{gom} & \textbf{mai} & \textbf{mar} & \textbf{nep} & \textbf{san} & \textbf{guj} & \multicolumn{1}{c}{\textbf{odi}} & \multicolumn{1}{c|}{\textbf{pan}} & \textbf{kan} & \textbf{mal} & \multicolumn{1}{c}{\textbf{tam}} & \multicolumn{1}{c|}{\textbf{tel}} & \textbf{mni}  & \textbf{brx}            & \textbf{sat}        \\
\cmidrule(lr){2-3}\cmidrule(lr){4-6}\cmidrule(lr){7-13}\cmidrule(lr){14-14}\cmidrule(lr){15-15}\cmidrule(lr){16-16}\cmidrule(lr){17-20}\cmidrule(lr){21-22}\cmidrule(lr){23-23}
\textbf{Language Script} & \multicolumn{2}{c}{Bangla}  & \multicolumn{3}{c}{Perso-Arabic}            & \multicolumn{7}{c}{Devanagari}                                                                         & Gujarati     & Odia                             & Gurmukhi                         & Kannada      & Malayalam    & Tamil                            & Telugu                            & Meitei        & Devanagari              & Ol Chik             \\ 
\textbf{No of Letters in Unicode}   & \multicolumn{2}{c}{96}      & \multicolumn{3}{c}{256}                     & \multicolumn{7}{c}{128}                                                                                 & 91           & \multicolumn{1}{c}{91}                               & \multicolumn{1}{c|}{80}                                & \multicolumn{1}{c}{91}           & \multicolumn{1}{c}{118}           & \multicolumn{1}{c}{72}                               & \multicolumn{1}{c|}{100}                                & 56            & \multicolumn{1}{c|}{96} & 48                  \\ 
\hdashline
\textbf{Models (Vocab)}   &&&&&&&&&&&&&&&&&&&&&& \\
\hdashline

\textbf{BLOOM (250680)}   & \multicolumn{2}{c}{(48,48)}      & \multicolumn{3}{c}{(49,207)}                     & \multicolumn{7}{c}{(67,61)}                                                                                 & (57,34)           & \multicolumn{1}{c}{(56,35)}                               & \multicolumn{1}{c|}{(55,25)}                                & \multicolumn{1}{c}{(62,29)}           & \multicolumn{1}{c}{(66,52)}           & \multicolumn{1}{c}{(46,26)}                               & \multicolumn{1}{c|}{(61,39)}                                & (00,56)            & \multicolumn{1}{c|}{(67,29)} & (00,48)                  \\

\textbf{FALCON (65024)}   & \multicolumn{2}{c}{(00,96)}      & \multicolumn{3}{c}{(12,244)}                     & \multicolumn{7}{c}{(2,126)}                                                                                 & (00,91)           & \multicolumn{1}{c}{(00,91)}                               & \multicolumn{1}{c|}{(00,72)}                                & \multicolumn{1}{c}{(0,100)}           & \multicolumn{1}{c}{(00,56)}           & \multicolumn{1}{c}{(02,70)}                               & \multicolumn{1}{c|}{(04,96)}                                & (00,56)            & \multicolumn{1}{c|}{(02,94)} & (00,48)                  \\
\textbf{LLAMA-1,2 (32024)}   & \multicolumn{2}{c}{(24,72)}      & \multicolumn{3}{c}{(45,211)}                     & \multicolumn{7}{c}{(38,90)}                                                                                 & (01,90)           & \multicolumn{1}{c}{(00,91)}                               & \multicolumn{1}{c|}{(04,76)}                                & \multicolumn{1}{c}{(02,89)}           & \multicolumn{1}{c}{(33,155)}           & \multicolumn{1}{c}{(19,53)}                               & \multicolumn{1}{c|}{(01,99)}                                & (00,56)            & \multicolumn{1}{c|}{(38,90)} & (00,48)                  \\
\textbf{MISTRAL (32052)}   & \multicolumn{2}{c}{(34,62)}      & \multicolumn{3}{c}{(47,209)}                     & \multicolumn{7}{c}{(43,85)}                                                                                 & (05,86)           & \multicolumn{1}{c}{(00,91)}                               & \multicolumn{1}{c|}{(02,78)}                                & \multicolumn{1}{c}{(18,73)}           & \multicolumn{1}{c}{(04,116)}           & \multicolumn{1}{c}{(22,50)}                               & \multicolumn{1}{c|}{(11,89)}                                & (00,56)            & \multicolumn{1}{c|}{(43,53)} & (00,48)                  \\
\textbf{MPT (50277)}   & \multicolumn{2}{c}{(05,91)}      & \multicolumn{3}{c}{(35,221)}                     & \multicolumn{7}{c}{(22,106)}                                                                                 & (02,89)           & \multicolumn{1}{c}{(00,91)}                               & \multicolumn{1}{c|}{(00,80)}                                & \multicolumn{1}{c}{(00,91)}           & \multicolumn{1}{c}{(01,117)}           & \multicolumn{1}{c}{(05,67)}                               & \multicolumn{1}{c|}{(03,97)}                                & (00,56)            & \multicolumn{1}{c|}{(22,106)} & (00,48)                  \\
\textbf{OPT (50265)}   & \multicolumn{2}{c}{(00,96)}      & \multicolumn{3}{c}{(13,243)}                     & \multicolumn{7}{c}{(1,127)}                                                                                 & (00,91)           & \multicolumn{1}{c}{(00,91)}                               & \multicolumn{1}{c|}{(00,80)}                                & \multicolumn{1}{c}{(00,91)}           & \multicolumn{1}{c}{(0,118)}           & \multicolumn{1}{c}{(00,72)}                               & \multicolumn{1}{c|}{(0,100)}                                & (00,56)            & \multicolumn{1}{c|}{(01,95)} & (00,48)                  \\
\bottomrule
\end{tabular}%
}
\caption{The language support of various LLMs for 22 Indian languages, along with the corresponding families, scripts, and letters representing each language. In each tuple (xx, yy), the first value represents the number of language-specific characters, while the second value indicates the count of byte-supported characters in respective LLM and for respective language.}
\label{tab:llms_details}
\end{table*}
%\input{llms_lang_support}
%\subsection{Indian Languages}
%In this study, we have considered a total of 22 scheduled Indian Languages, which can be categorized into four main language families: Indo-Aryan, Dravidian, Sino-Tibetan, and Austroasiatic. Writing systems play a crucial role in representing these languages, and in this context, we have 13 major writing scripts that are used to represent these languages. It is interesting to note that most of these scripts can be traced back to the Brahmi script, which served as the foundation for the development of several Indian scripts. Each of these 13 writing systems has its own unique set of letters and characters, reflecting the phonetic and linguistic characteristics of the respective languages they represent. For any corpus, by studying the available letters (or characters) one can get valuable insights for representatives and coverage for any language scripts.

Pre-trained (or Raw) large language models are trained on a huge amount of language data, and some of the these models are trained on multiple languages \citep{naveed2023comprehensive}. However, their training primarily focuses on English text \citep{penedo2023refinedweb}. Emphasis on English is due to its substantial presence on the internet and its widespread usage in business contexts. For the purpose of this work, our objective is to assess the effectiveness of these models in Machine Translation tasks that involve both English and Indian Languages. Consequently, it becomes crucial to investigate the representation of Indian languages within these large language models.\\

%An approach to investigate representation of Indian languages within large language model can be by analyzing the frequency of language-specific words and sentences utilized during the training of these models. Unfortunately, it is not possible to perform this analysis as the training data used for these models is not accessible to the public. LLaMA-2 has mentioned that its pre-training corpus primarily consists of English and may not be optimal for other languages \citep{touvron2023LLaMA2}. However, it is worth mentioning that approximately 8.38\% of the data does include other languages then English and codes in LLaMA-2.

%An approach to investigating the representation of Indian languages within a large language model can involve analyzing the frequency of language-specific words and sentences used during the training of these models. Unfortunately, it is not possible to perform this analysis as the training data used for these models is not publicly accessible. LLaMA-2, in particular, has mentioned that its pre-training corpus primarily consists of English and may not be optimal for other languages \citep{touvron2023LLaMA2}. However, it is worth mentioning that approximately 8.38\% of the data does include languages other than English and codes in LLaMA-2.

An approach to investigating the representation of Indian languages within a large language model can involve analyzing the frequency of language-specific words and sentences used during the training of these models. Unfortunately, it is not possible to perform this analysis as the training data used for these models are not publicly accessible. LLaMA-2, in particular, has mentioned that its pre-training corpus primarily consists of English and may not be optimal for other languages \citep{touvron2023LLaMA2}. However, it is worth mentioning that approximately 8.38\% of the data does include languages other than English and codes in LLaMA-2.

%On other hand, by studying the vocabulary (or letters/characters) of a corpus, valuable insights can be gained regarding the representation and coverage of language in that corpus. Writing system or script plays a crucial role in representing language. Hence, analysis of vocabulary can be considered as a proximal task. Any we have access to the sub-word vocabulary for considered LLMs. By tallying the characters present in the sub-word vocabulary and in the corresponding language script, we can approximate the language representation within the respective large language model. In this work, we included a total of 22 scheduled Indian Languages for translation, which can be categorized into four main language families: Indo-Aryan, Dravidian, Sino-Tibetan, and Austroasiatic. 13 major scripts are used for these 22 Indian languages in writing system. It is interesting to note that most of these scripts can be traced back to the Brahmi script \footnote{\url{https://en.wikipedia.org/wiki/Brahmi_script}}, which served as the foundation for the development of several Indian scripts \citep{f8a942cf-ee25-36c9-a48e-4b71549fddab}. Each of these 13 writing systems \footnote{\url{https://en.wikipedia.org/wiki/Official_scripts_of_the_Republic_of_India}} has its own unique set of letters and characters, reflecting the phonetic and linguistic characteristics of the respective languages they represent. \\

On the other hand, studying the vocabulary (or letters/characters) of a corpus can provide valuable insights into the representation and coverage of language within that corpus. The writing system or script used plays a crucial role in representing a language. Therefore, analyzing the vocabulary can be considered a proximal task. Fortunately, we have access to the sub-word vocabulary for the considered large language models. By comparing the characters present in the sub-word vocabulary with those in the corresponding language script, we can approximate the language representation within the respective LLM.\\

For this work, we included a total of 22 scheduled Indian languages for translation, which can be categorized into four main language families: Indo-Aryan, Dravidian, Sino-Tibetan, and Austroasiatic. These 22 Indian languages are written using 13 major scripts. It is interesting to note that most of these scripts can be traced back to the Brahmi script \footnote{\url{https://www.education.gov.in/sites/upload_files/mhrd/files/upload_document/languagebr.pdf}}, which served as the foundation for the development of several Indian scripts \citep{f8a942cf-ee25-36c9-a48e-4b71549fddab}. Each of these 13 writing systems has its own unique set of letters and characters \footnote{\url{https://en.wikipedia.org/wiki/Official_scripts_of_the_Republic_of_India}}, reflecting the phonetic and linguistic characteristics of the respective languages they represent.\\

%Table~\ref{tab:llms_details} presents an overview of these scripts, the languages utilizing these scripts, and the corresponding sub-word vocabulary sizes for these languages in their respective LLMs. The numbers indicated in `(X,Y)' represent the counts of native script characters present in the respective LLM. Specifically, X denotes the number of native language characters indexed in the vocabulary, while Y denotes the number of characters represented as pre-defined hexadecimal values (UNKs). Upon analysis, we observe that, in general, 22 Indian languages have a limited presence in the most of the LLMs'. The Devanagari, Perso-arabic and Bangla scripts demonstrate the most extensive sub-word vocabularies, whereas other scripts exhibit minimal or near-zero representation within the vocabulary.

Table~\ref{tab:llms_details} presents an overview of the scripts, the languages utilizing these scripts, and the corresponding sub-word vocabulary sizes for LLMs. The numbers indicated in `(X,Y)' represent the counts of native script letters (characters in unicode \footnote{\url{https://unicode.org/}}) present and not present in the respective LLM. Specifically, X denotes the number of native language characters present in the vocabulary, while Y denotes the number of characters represented as pre-defined (multiple) hexadecimal values. Upon analysis, we observe that, in general, the 22 Indian languages have a limited presence in most of the LLMs. However, the Devanagari, Perso-Arabic, and Bangla scripts demonstrate the most extensive sub-word vocabularies, while other scripts have minimal or near-zero representation within the vocabulary.

\begin{figure*}
    \centering
    \includegraphics[width=.9\textwidth]{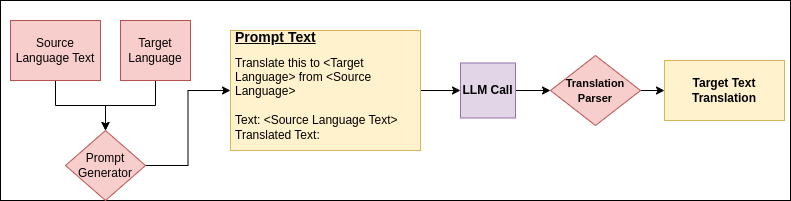}
    \caption{Prompting Mechanism for Translation}
    \label{fig:prompting-1}
\end{figure*}

\section{Experiment setup: Machine Translation using LLMs}

%To assess the effectiveness of the above LLMs for machine translation tasks involving English and 22 Indian languages, we conducted two experiments. In the first experiment, we evaluated the performance of the pre-trained (raw) LLM. In the second experiment, we employed example-based in-context learning for the same machine translation task. Both translation directions were explored, namely English to 22 Indian languages and 22 Indian languages to English. All the experiments are carried out on translation benchmark data as mentioned in Section \ref{MBD}.\\

To evaluate the performance of the large language models (LLMs) in machine translation tasks involving English and 22 Indian languages, we conducted two experiments. The first experiment focused on assessing the performance of the pre-trained (raw) LLM. In the second experiment, we utilized example-based in-context learning for the same machine translation task. Both translation directions were explored, including English to 22 Indian languages and 22 Indian languages to English. All experiments were conducted using translation benchmark data, as discussed in Section \ref{MBD}.\\

As part of our experimental setup, we used the prompting pipeline depicted in Figure \ref{fig:prompting-1}. This pipeline involved using a Prompt Generator to generate specific prompts for the source and target language along with source text. Subsequently, an LLM call is triggered to generate a response, which was then processed by a translation parser to obtain the actual translation. To ensure high-throughput and memory-efficient inference and serving for LLMs, we utilized the vLLM library\footnote{\url{https://github.com/vllm-project/vllm}} \citep{kwon2023efficient}. We conducted all experiments using a temperature parameter of 0, which ensures that the model behaves deterministically. By setting the temperature to 0, the model is constrained to select the word with the highest probability, effectively limiting its choice to the most likely option \citep{aksitov2023characterizing}. All of our experiments are conducted using vLLM library on A100, 40GB GPUs.

\begin{table*}[!ht]
    \centering
    
    \begin{tabular}{l | r r r r r r r r r}
       \toprule
       \textbf{English-}& \textbf{\#Sents} & \textbf{S-AvgL}& \textbf{T-AvgL} &\textbf{S-Words} & \textbf{T-Words} & \textbf{S-Types} & \textbf{T-Types}\\ \hline
       \textit{Assamese (asm)} &  138208 & 16.88 & 14.39 & 2333583 & 1988395 & 125480 & 185151 \\ 
       \textit{Bangla (ban)} & 180219 & 17.80 & 15.07 & 3208203 & 2715959 & 161820 & 227468\\ 
       \textit{Bodo (brx)} &113139 & 17.79 & 13.96 & 2012274 & 1579042 & 116963 & 227180\\ 
       \textit{Dogri (doi)} & 24157 & 15.32 & 17.68 & 370047 & 427110 & 48256 & 41370\\ 
       \textit{Konkani (gom)} & 97555 & 17.13 & 14.03 & 1671465 & 1368512 & 82783 & 145300\\ 
       \textit{Gujarati (guj)} & 135664 & 17.71 & 15.96 & 2402552 & 2164831 & 123935 & 174886\\
       \textit{Hindi (hin)} & 222356 & 17.84 & 19.69 & 3966247 & 4378231 & 183737 & 202423\\ 
       \textit{Kannada (kan)} & 117222 & 16.83 & 12.44 & 1972881 & 1458053 & 100778 & 208803\\ 
       \textit{Kashmiri (kas)} & 19824 & 16.02 & 17.68 & 317634 & 350577 & 43197 & 66210\\ 
       \textit{Maithili (mai)} &  23690 & 16.11 & 15.79 & 381720 & 374042 & 52920 & 57423\\ 
       \textit{Malayalam (mal)} & 137950 & 16.30 & 11.13 & 2248081 & 1535654 & 120999 & 299146\\ 
       \textit{Marathi (mar)} &  175893 & 17.94 & 14.81 & 3154904 & 2604119 & 167822 & 299983\\ 
       \textit{Meitei (mni)} & 56617 & 17.77 & 15.73 & 1006271 & 890828 & 86175 & 161043\\ 
       \textit{Nepali (nep)} & 85442 & 16.76 & 14.13 & 1431858 & 1207687 & 105411 & 145175\\
       \textit{Odia (odi)} & 36923 & 17.07 & 15.49 & 630148 & 571958 & 68765 & 79932\\ 
       \textit{Punjabi (pan)} & 80951 & 17.22 & 18.29 & 1394286 & 1480835 & 63510 & 74451\\
       \textit{Sanskrit (san)} & 33189 & 16.30 & 11.69 & 541034 & 387957 & 61591 & 119856\\
       \textit{Santali (sat)} &24368 & 16.95 & 19.28 & 412918 & 469791 & 51307 & 56053\\
       \textit{Sindhi (sin)} & 10503 & 17.10 & 19.32 & 179592 & 202952 & 28945 & 30782\\
       \textit{Tamil (tam)} & 150254 & 17.76 & 13.34 & 2668252 & 2004981 & 139214 & 290917\\
       \textit{Telugu (tel)}&  111808 & 16.81 & 12.64 & 1879737 & 1413466 & 96105 & 191792\\
       \textit{Urdu (urd)} & 150747 & 17.62 & 20.20 & 2656814 & 3044480 & 144001 & 129856\\
       \bottomrule
    \end{tabular}
    \caption{English to Indian Languages machine translation Fine-tuning data from BPCC-Human \citep{ai4bharat2023indictrans2}. In this, the term "\#Sents" refers to the total number of parallel sentences. "S-AvgL" and "T-AvgL" represent the average sentence length, in terms of words, for the source and target languages, respectively. Likewise, "Words" denotes the total number of words, while "Type" represents the total number of unique words.}
    \label{tab:DevelopmentDataforTranslation}
\end{table*}
\subsection{Machine Translation on Raw LLM}
\label{LLMPromp1}
\label{LLMPromp2}
To optimize the machine translation task on our selected LLMs, we conducted manual trials with various prompts. Through these trials, we discovered that directly asking for the translation and presenting the text in JSON format yielded better results, as the models seemed to comprehend the JSON structure more effectively \citep{reinauer2023neural}. After multiple iterations, we finalized two prompts for translating sentences using raw (pre-trained) LLMs, as illustrated in below examples. These prompts were used to evaluate the efficiency of the models. 

\myexample{Translation Prompt-1}{
\label{prompt1-ex}
Translate this to <Target Language> from <Source Language>\\

Text: <Source Language Text>\\
Translated Text: 
}

\myexample{Translation Prompt-2}{
\label{prompt2-ex}
Translate this from <Source Language> to <Target Language>\\

<Source Language>: <Source Language Text>\\
<Target Language>: \\
}
\begin{center}
\label{prompt3-ex}
\myexample{ICL Translation Prompt}{
If the <Source Language> to <Target Language> translation for "<Source Example>" is "<Target Example>" from <Source Language>, following that, translate this to <Target Language> from <Source Language> \\

Text: <Source Language Text>\\
Translated Text: 
}    
\end{center}

Similarly, we identified and modified the prompt for example-based in-context learning with LLM. This prompt is specified in Example above (ICL Translation Prompt). In the case of in-context learning, all of our experiments involved providing a single translation sample as a contextual learning example prior to the actual translation command. We ensured that this example remained consistent for the same language pair across the sentences. The sample itself was randomly selected from the Human-BPCC translation training corpus \citep{ai4bharat2023indictrans2}. 
We present the outcomes of both of these experiments in the Performance and Discussions section.

\subsection{Fine-tuning LLM for Machine Translation}
To examine the potential improvement in multilingual understanding or translation performance of LLMs beyond the pre-trained LLM baseline, we conducted fine-tuning experiments for the translation task. 

\subsubsection{Training Data}
\label{trainingdata}
To fine-tune large language models (LLMs) for the machine translation task, we utilized the Bharat Parallel Corpus Collection (BPCC). This corpus is publicly available and specifically for English to 22 Indic languages translation. It consists of two main parts: BPCC-Mined and BPCC-Human, comprising a total of approximately 230 million parallel text pairs. For the fine-tuning process, we focused on the BPCC-Human dataset, which contains 2.2 million English-Indic pairs. Additionally, this dataset includes subsets derived from English Wikipedia sentences and everyday usage scenarios. For more information about this corpus, are presented in Table \ref{tab:DevelopmentDataforTranslation}.\\

\begin{table}
\centering
\resizebox{\columnwidth}{!}{%
\begin{tabular}{lll}
\toprule
\textbf{Method}                          & \textbf{Hyper-parameter}    & \textbf{Value}  \\
\midrule
\multirow{7}{*}{LoRA}           & LoRA modules      &      PEFT\footnote{\url{https://github.com/huggingface/peft}}  \\
                                & rank              &    8    \\
                                & dropout           &    0.05   \\
                                & learning rate     &    1e-4    \\
                                & global batch size &     8   \\
                                & epochs            &     6   \\
\midrule
\multirow{3}{*}{Full-parameter FSDP} & learning rate     &    1e-4    \\
                                & global batch size &    4    \\
                                & epochs            &    5   \\
\bottomrule
\end{tabular}
}
\caption{Hyper-parameter configurations of LoRA based and full fine-tuning for 4*A100 40GB GPUs}
\label{LLM:Hyper-parameter}
\end{table}

% \begin{table}
% \centering
% \begin{tabular}{lll}
% \toprule
% \textbf{Method}                          & \textbf{Hyper-para}    & \textbf{Value}  \\
% \midrule
% \multirow{6}{*}{LoRA/Full}           & LoRA      &     PEFT\footnote{https://github.com/huggingface/peft}; FSDP\footnote{for Full Fine-Tuning}   \\
%                                 & rank              &    8    \\
%                                 & dropout           &    0.05    \\
%                                 & learning rate     &    1e-4    \\
%                                 & batch size &      4  \\
%                                 & epochs            &      5  \\
% \bottomrule
% \end{tabular}
% \caption{Hyper-parameter configurations of LoRA based and full fine-tuning for 4*A100 40GB GPUs}
% \label{LLM:Hyper-parameter}
% \end{table}
% Please add the following required packages to your document preamble:
% \usepackage{multirow}
% \usepackage{graphicx}
\begin{table*}[!htb]
\centering
\resizebox{2.05\columnwidth}{!}{%
\begin{tabular}{|l|l|l|}
\hline
\textbf{TestSet}  & \textbf{\#Sent} & \textbf{Details}                                                                                                                                                        \\ \hline
IN22\_conv\_test  & 1502            & \multirow{2}{*}{\citeauthor{ai4bharat2023indictrans2} released MT benchmark data covering English to 22 Indian Languages.}                             \\ \cline{1-2}
IN22\_gen\_test   & 1023            &                                                                                                                                                                         \\ \hline
Flores200-dev     & 997             & \multirow{2}{*}{\citeauthor{goyal-etal-2022-flores} released MT benchmark data which includes English to 17 Indian Language pairs considered in this work.} \\ \cline{1-2}
Flores200-devtest & 1012            &                                                                                                                                                                         \\ \hline
Newstest2019      & 1997            & \citeauthor{federmann-etal-2022-ntrex}  released MT benchmark data which includes English to 10 Indian Language pairs considered in this work.              \\ \hline
\end{tabular}%
}
\caption{Benchmark data details covering English to 22 Indian Languages}
\label{tab:testdata}
\end{table*}
\begin{figure*}
    \centering
    \includegraphics[width=.5\textwidth]{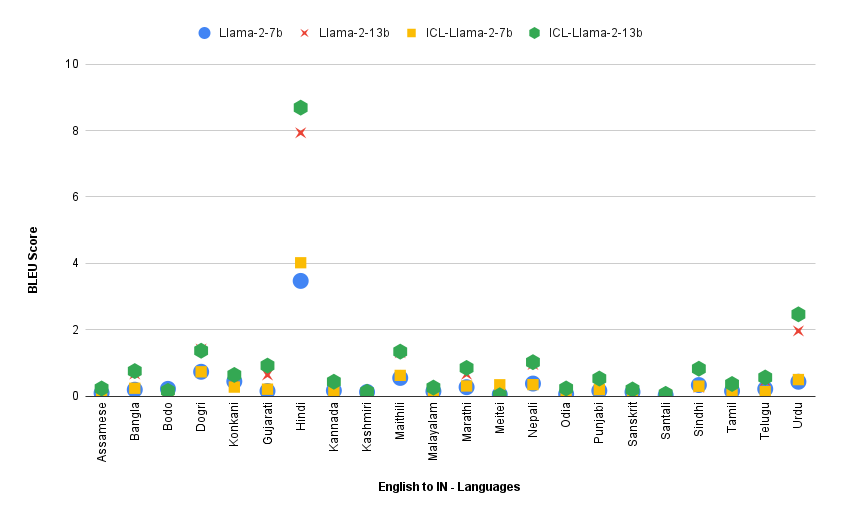}\hfill
    \includegraphics[width=.5\textwidth]{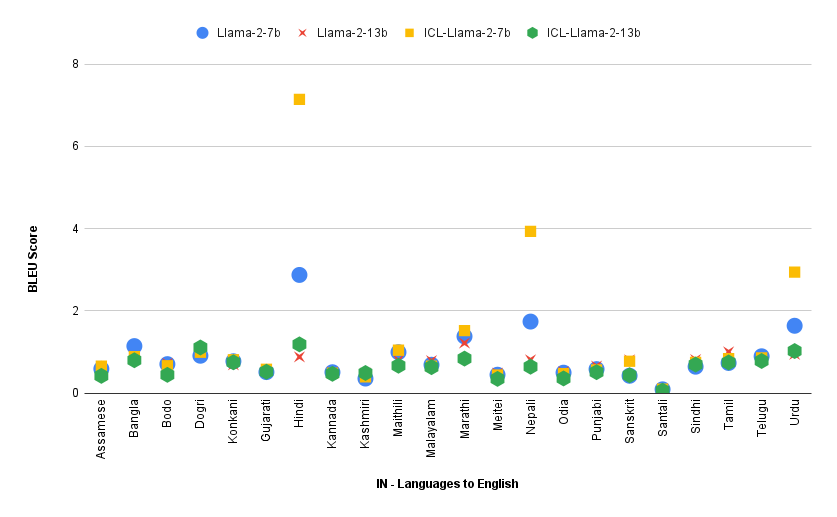}\hfill
    \caption{Evaluation of English - 22 Indic language Translation over 5 benchmark-sets (averaged): Raw LLM vs In Context Learning (ICL); Raw LLM models: LLaMA-2-7b, LLaMA-2-13b)}
    \label{fig:RawvsICLResults}
\end{figure*}
\subsubsection{Fine-tuning Details}
Considering the raw LLM performance, model parameters, and resource constraints, we selected a subset of LLMs for the fine-tuning process. Specifically, we chose LLaMA-2-7b, LLaMA-2-13b, and Mistral-7B for the fine-tuning experiment. For the selected LLMs, we decided to conduct fine-tuning using multiple parameters to enhance their performance. These parameters included bi-lingual translation fine-tuning, multi-lingual translation fine-tuning, low-rank adaptation-based fine-tuning, and a two-stage fine-tuning approach: full fine-tuning followed by low-rank adaptation-based fine-tuning. Due to limitations in training resources, we prioritized full fine-tuning as the chosen option.\\

Specifically, we performed LoRa-based fine-tuning \citep{hu2021lora} for all English to 22 Indian languages (in both directions) under bi-lingual settings using LLaMA-2-7b and LLaMA-2-13b. Additionally, we conducted multi-lingual LoRa-based fine-tuning for English to the combined 22 Indian languages, as well as for the combined 22 Indian languages to English, using LLaMA-2-7b, LLaMA-2-13b, and Mistral-7B. Based on the overall performance, we proceeded with a two-stage fine-tuning approach for the multi-lingual translation task specifically on LLaMA-2-13b. In the first stage, we performed full fine-tuning as a multi-lingual translation setup. Subsequently, in the second stage, we conducted multi-lingual LoRa-based fine-tuning on the same fully fine-tuned model.\\

For both types of fine-tuning LLMs, we utilized the llama-recipes codebase\footnote{\url{https://github.com/facebookresearch/llama-recipes/}} which provides an efficient implementation for LoRa-based adaptor fine-tuning with PEFT \citep{peft}. For more details, please refer to the llama-recipes documentation \footnote{\url{https://github.com/facebookresearch/llama-recipes/blob/main/docs/LLM_finetuning.md}}. The hyperparameters for the fine-tuning process are specified in Table \ref{LLM:Hyper-parameter}. The training data used for the fine-tuning experiments will be presented in the sub-section \ref{trainingdata}.

\section{Machine Translation Benchmark Data}
\label{MBD}
We evaluate the performance of multilingual translation using three different benchmark datasets, as outlined in Table \ref{tab:testdata}. The table provides a comprehensive overview of each dataset, highlighting the availability of n-way parallel data for the specified number of Indian languages from English as a source direction.

\label{LLMPromp3}
\label{LLMPromp4}
\begin{figure*}
    \centering
    \includegraphics[width=.5\textwidth]{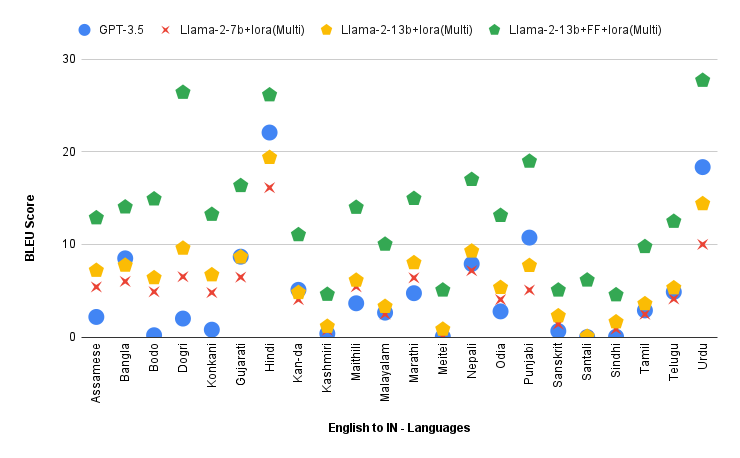}\hfill
    \includegraphics[width=.5\textwidth]{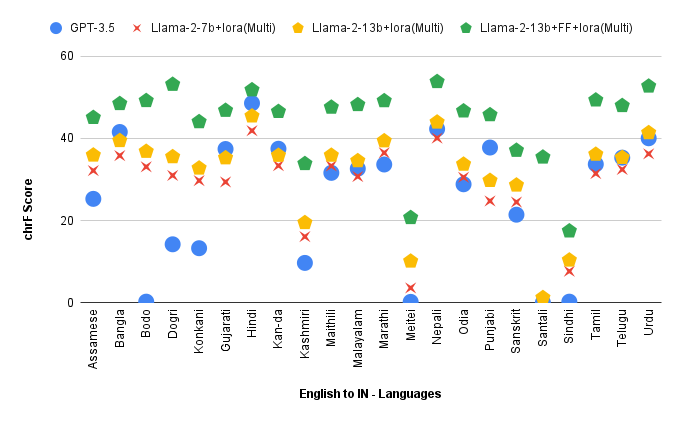}\hfill
    \caption{Performance comparison of GPT-3.5 vs our Fine-Tuned LLM Translation models (LLaMA-2-7b+lora (Multi), LLaMA-2-13b+lora (Multi), and LLaMA-2-13b+FF+lora (Multi)): English to 22 Indian Languages over 5 benchmark-sets (averaged). Here, LORA stands for Low-Rank Adaptation of Large Language Models based fine-tuning. Multi stands for the multilingual model, FF for full-finetuning, and FF+lora stands for 2-stage fine-tuning.}
    \label{fig:BLEUCHRFEN2XX}
\end{figure*}

\section{Performance Evaluation}
We evaluated the performance of the translation outputs using BLEU \citep{papineni2002bleu} and chrF \citep{popovic2015chrf} evaluation methods on benchmark data described in Section~\ref{MBD}. However, we did not include COMET \citep{rei-etal-2022-comet} as an evaluation method due to the absence of support for many low-resource Indian languages at the time of evaluation. We used sacreBLEU library \citep{post-2018-call} for BLEU \footnote{footprint for BLEU: \\nrefs:1|case:mixed|eff:no|tok:13a|smooth:exp|version:2.1.0} and chrF \footnote{footprint for chrF: \\ nrefs:1|case:mixed|eff:yes|nc:6|nw:0|space:no|version:2.1.0} calculation. To mitigate the impact of randomness in scores, we present our findings as the average of two runs for all of our results. 

\paragraph{Raw (Zero shot) vs ICL based Translation on LLMs}

Figure \ref{fig:RawvsICLResults} presents the comparison of overall results when evaluating translation quality for Raw LLMs and In Context Learning (ICL) based LLMs outputs. The left sub-figure represents the results for English to 22 Indian languages, while the right sub-figure presents the results for 22 Indian languages to English translation. We observed amplified performance for the Bloom large language model for certain languages, which can be attributed to the known MT benchmark data leak in the pre-training \citep{zhu2023multilingual}. Consequently, we decided to exclude this language model from further experiments. \\

LLMs models such as OPT, MPT, LLAMA-1 and Falcon exhibited poor performance, which can be correlated with the no or minimal presence of characters of our focused Indian Languages in their vocabulary (Table \ref{tab:llms_details}). Therefore, we have omitted reporting the results for these models. Figure \ref{fig:RawvsICLResults} indicates that the Llama-2 models show relatively better performance with ICL settings compared to the raw models. Detailed results are presented in appendix.\\

Through manual analysis, we observed that for less-represented languages such as Gujarati, Kannada, Odia, etc. (Table \ref{tab:llms_details}), the ICL-driven translation tends to repeat the same translation given in the context as learning. On the other hand, the raw models tend to hallucinate and repeat words throughout the translation \citep{guerreiro2023hallucinations}. \\

One important finding from the manual analysis is that these raw LLMs demonstrate the ability to accurately identify languages (e.g., when asked for Gujarati translation, it gives inaccurate translations but correctly hallucinate text in the Gujarati script). This is a positive aspect and indicates a significant advantage of these LLMs in terms of their understanding and differentiation of languages and language scripts. In response to the question asked in Introduction, it is true that the major available LLMs are primarily focused on English. However, \textit{they do exhibit minimal potential for zero-shot and example-based translation capabilities}.\\

\paragraph{Fine-Tuned LLM driven Translations: English to Indian Languages}
\begin{figure*}
    \centering
    \includegraphics[width=.5\textwidth]{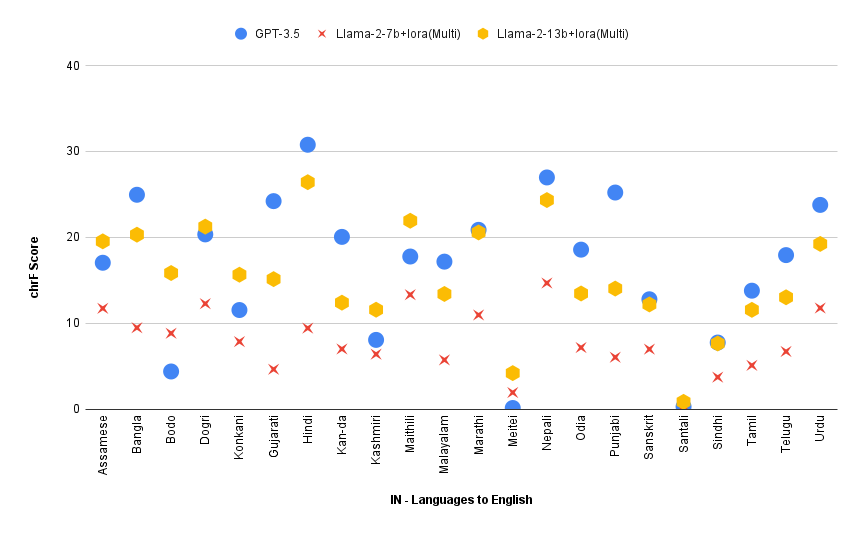}\hfill
    \includegraphics[width=.5\textwidth]{NewXXEnbleu.png}\hfill
    \caption{Performance comparison of GPT-3.5 vs our Fine-Tuned LLM Translation models (LLaMA-2-7b+lora (Multi), LLaMA-2-13b+lora (Multi), and LLaMA-2-13b+FF+lora (Multi)): English to 22 Indian Languages over 5 benchmark-sets (averaged). Here, LORA stands for Low-Rank Adaptation of Large Language Models based fine-tuning. Multi stands for the multilingual model.}
    \label{fig:BLEUCHRFXX2EN}
\end{figure*}
We conducted an evaluation to compare the performance of our Fine-Tuned LLM models with GPT-3.5, as both models use the same decoder-based approach. Figure \ref{fig:BLEUCHRFEN2XX} illustrates the comparison for English to 22 Indian language translation. The scores for GPT-3.5 are generally lower compared to our fine-tuned methods, also our fine-tuned models have higher numbers than our previously mentioned zero-shot and example-based learning baseline. This indicates that with minimal translation corpora, we are able to achieve considerable translations for translating into Indian languages from English.\\

Additionally, we observed that multilingual fine-tuning yielded better overall performance compared to bilingual fine-tuning. The two-stage fine-tuning approach also outperformed other fine-tuning methods for the translation task. The impressive results of the two-stage fine-tuning approach, as shown in Figure \ref{fig:BLEUCHRFEN2XX}, are comparable to those of traditional encoder-decoder based translation models. It is worth noting that this performance improvement was achieved using only a few thousand parallel data, whereas traditional NMT models typically require a larger amount of data. From Figure \ref{fig:BLEUCHRFEN2XX}, we can see that translating to low-resource languages such as Dogri, Konkani, Kashmiri, Meitei, Sanskrit, and Sindhi yielded favorable evaluation numbers (Detailed results are presented in appendix) compared to existing translation systems. In answer to the question posed in the introduction, \textit{fine-tuning LLMs does enhance translation capabilities, particularly more when employing multilingual fine-tuning. These models demonstrate proficiency in translating low-resource languages as well.}\\

\paragraph{Fine-Tuned LLM driven Translations: Indian Languages to English}
Figure \ref{fig:BLEUCHRFXX2EN} showcases the comparison for Indian language to English translation. The scores for GPT-3.5 are generally not higher compared to our fine-tuned methods, while our fine-tuned models still outperform the previously mentioned zero-shot and example-based context learning driven LLM results. Notably, the performance improvement for Indian language to English translation is comparatively lower than that of English to Indian language translation. Compared to translations from English to Indian languages, the LoRa-based single-stage fine-tuning here performs the best among all the fine-tuning approaches. Detailed results are presented in the appendix.\\

This disparity can be attributed to the vocabulary representation of Indian languages in these LLMs. As presented in Table \ref{tab:llms_details}, the subword vocabulary for Indian languages is limited in the considered LLMs. Consequently, when processing input in Indian languages, characters that are not present in the vocabulary receive multiple hexadecimal representations from the vocabulary. This creates a bottleneck in understanding the underlying meaning, making it challenging for the larger LLM network to establish corresponding semantic translations.\\

However, this issue does not arise when translating from English to Indian languages. The underlying understanding of English is robust, allowing the network to effectively map the respective language translations.\\

Hence, this suggest the need for LLMs where enough language representation is required and future development of LLMs must address this.

\section{Limitations}
In order to conduct our experiments, we relied on high-performance GPUs, specifically the A100-40GB. However, we acknowledge that not everyone may have access to such powerful computing resources, making it challenging to reproduce our experiments and achieve identical results. To overcome this limitation, our objective is to provide open access to all outputs, including model and results, to facilitate further research and exploration. By making these resources openly available, we aim to promote collaboration and enable others to build upon our work.

\section{Conclusion}

Our experiments and results have provided promising insights into the use of LLMs for translation tasks. We have found that LLMs have the potential to perform translations involving English and Indian languages without the need for an extensive collection of parallel data, which distinguishes them from traditional translation models. Furthermore, our findings indicate that LLaMA-2 based models outperform other models in zero-shot and in-context example-based learning. Notably, the LLaMA-2-13b based model demonstrates superior performance compared to its counterparts. To enhance the LLM's understanding of English and Indian languages, we have introduced a two-stage fine-tuning process. This process begins with initial full-finetuning, followed by LoRa-based fine-tuning. Through this approach, we have significantly improved the LLM's comprehension of content in both languages.\\

However, our experiments suggest that further work on LLMs is required to surpass the performance of traditional encoder-decoder based translation models. This work could involve the development of Indian language-specific LLMs, which would enhance vocabulary and alphabet coverage, resulting in better representation of Indian languages.\\

On the other hand, in the future, we plan to incorporate Indian to Indian language translation using LLMs. Additionally, our aim is to develop a single LLM model capable of translating all Indian languages, as well as English, in both directions. By doing so, we strive to push the boundaries of language capabilities within LLMs and further advance the field.

\section*{Acknowledgement}

We express our gratitude to Pruthwik Mishra, Arafat Ahsan and Palash Gupta for their contributions throughout the different phases of this project. This undertaking is funded by the Ministry of Electronics and Information Technology, Government of India, as evidenced by the Sanction Order: 11(1)/2022-HCC(TDIL)-Part(2)/A/B/C and the Administrative Approval: 11(1)/2022-HCC(TDIL)-Part(2).

% \bibliography{anthology,custom}
\bibliography{custom} %modified temporarly

\newpage
\onecolumn
%\newpage
\appendix
\section{Appendix}
\label{sec:appendix}
Examples 

\begin{figure*}[!ht]
    \centering
    \includegraphics[width=.9\textwidth]{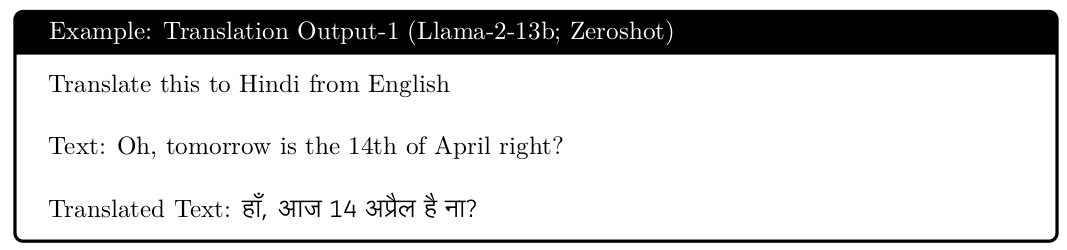}
    \label{fig:enter-label}
\end{figure*}

\begin{figure*}[!ht]
    \centering
    \includegraphics[width=.9\textwidth]{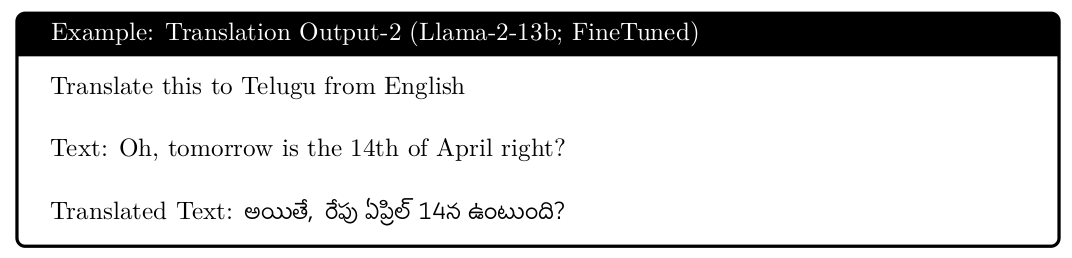}
    \label{fig:enter-label}
\end{figure*}

\begin{figure*}[!ht]
    \centering
    \includegraphics[width=.9\textwidth]{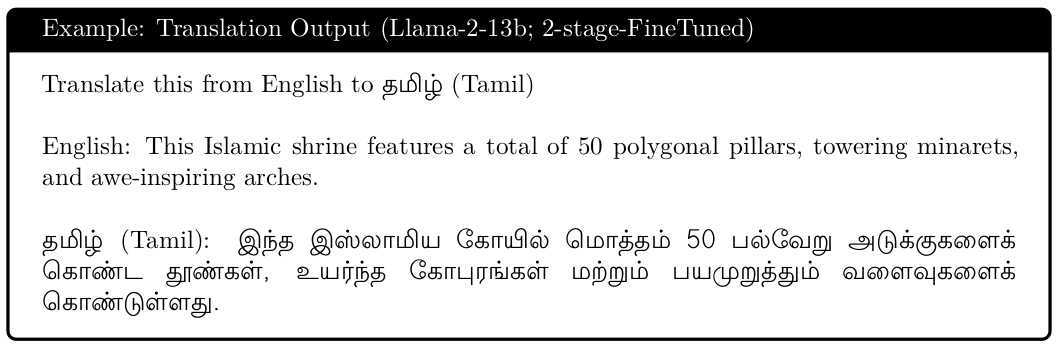}
    \label{fig:enter-label}
\end{figure*}

\begin{figure*}[!ht]
    \centering
    \includegraphics[width=.9\textwidth]{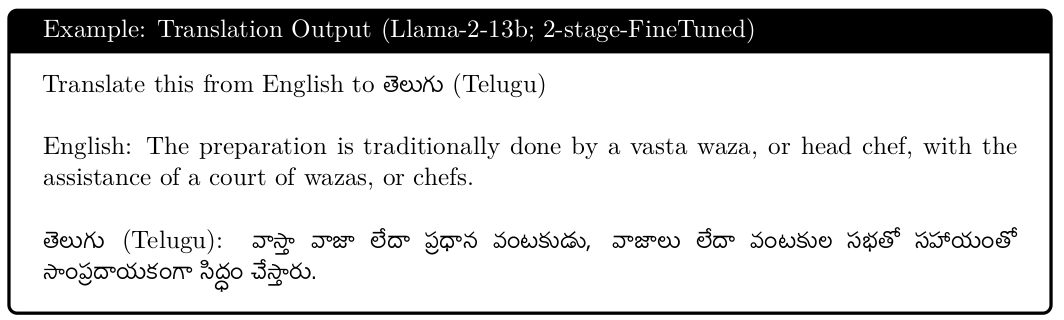}
    \label{fig:enter-label}
\end{figure*}

\begin{figure*}[!ht]
    \centering
    \includegraphics[width=.9\textwidth]{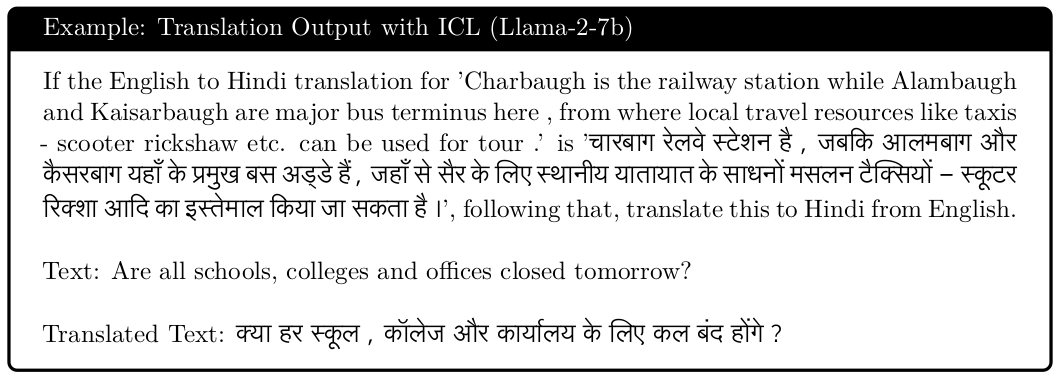}
    \label{fig:enter-label}
\end{figure*}

\begin{figure*}[!ht]
    \centering
    \includegraphics[width=.9\textwidth]{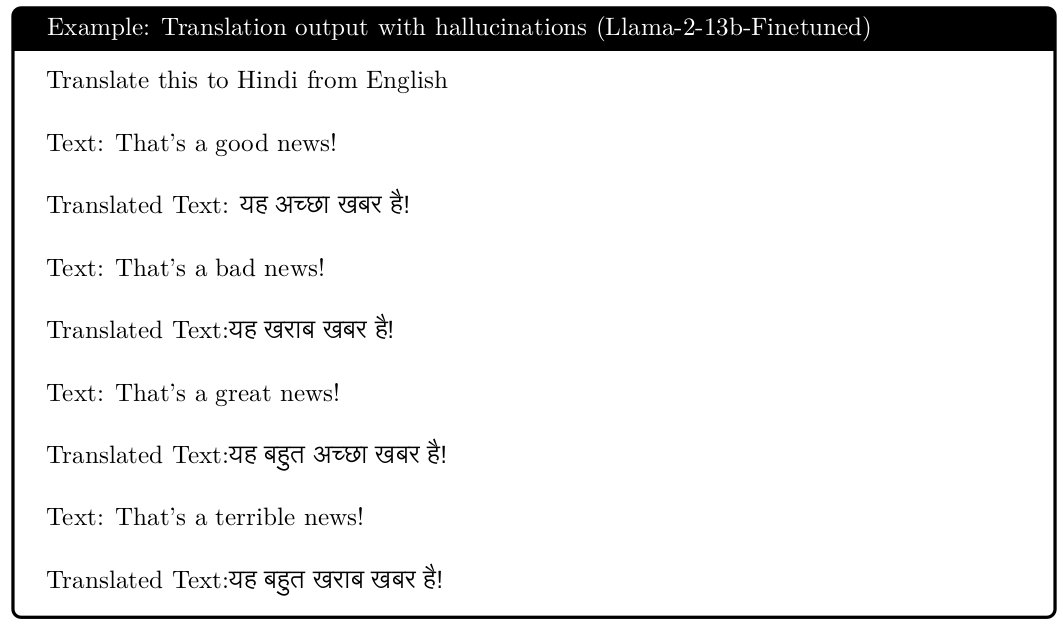}
    \label{fig:enter-label}
\end{figure*}
\begin{table*}[h]
\centering
\resizebox{\columnwidth}{!}{%
\begin{tabular}{llllllllllllllllllllllll}
\toprule
\textbf{DataSet} & \multicolumn{1}{l}{\textbf{Model}} & \multicolumn{1}{l}{\textbf{asm}} & \multicolumn{1}{l}{\textbf{ban}} & \multicolumn{1}{l}{\textbf{bod}} & \multicolumn{1}{l}{\textbf{doi}} & \multicolumn{1}{l}{\textbf{kon}} & \multicolumn{1}{l}{\textbf{guj}} & \multicolumn{1}{l}{\textbf{hin}} & \multicolumn{1}{l}{\textbf{kan}} & \multicolumn{1}{l}{\textbf{kas}} & \multicolumn{1}{l}{\textbf{mai}} & \multicolumn{1}{l}{\textbf{mal}} & \multicolumn{1}{l}{\textbf{mar}} & \multicolumn{1}{l}{\textbf{mei}} & \multicolumn{1}{l}{\textbf{nep}} & \multicolumn{1}{l}{\textbf{odi}} & \multicolumn{1}{l}{\textbf{pun}} & \multicolumn{1}{l}{\textbf{san}} & \multicolumn{1}{l}{\textbf{sat}} & \multicolumn{1}{l}{\textbf{sin}} & \multicolumn{1}{l}{\textbf{tam}} & \multicolumn{1}{l}{\textbf{tel}} & \textbf{urd} \\ \midrule
\multirow{11}{*}{\textit{\textbf{IN22\_conv}}} & GPT-3.5 & 2.4 & 9.6 & - & 1.2 & 0.2 & 11.1 & 22.3 & 2.6 & 0.1 & 1.6 & 1.6 & 5.7 & 0.1 & 9.5 & 2.3 & 12.3 & 0.5 & - & - & 2.8 & 4.7 & 21.9 \\
 & IndicTrans-2 & 15.9 & \textbf{16.6} & 12 & \textbf{26.1} & \textbf{13.4} & \textbf{26.8} & 27.6 & 5.4 & 2.7 & \textbf{17.2} & \textbf{5.5} & \textbf{18.8} & 7.1 & 19.4 & 9.4 & 30 & 5.4 & 6.3 & 5.2 & \textbf{7.4} & \textbf{13.5} & \textbf{38.4} \\
 & Google Translate & 13.9 & - & - & 14.3 & 11.9 & 26.6 & \textbf{28.8} & 5.2 & - & 9.2 & \textbf{5.5} & 17.6 & - & 14.5 & 9.3 & - & - & - & - & 8 & 13.1 & 37.1 \\
 & LTRC, IIIT-H & - & - & - & - & - & 17.1 & 22.5 & 3.4 & - & - & 3.5 & 11.9 & - & - & - & - & - & - & - & 5.7 & 10.2 & 21.8 \\
 & SeamlessM4T & \textbf{16.2} & 15.6 & 0 & 0 & 0 & - & 24.5 & 4.7 & 0 & 15.4 & \textbf{5.5} & 18 & 0 & 15.7 & \textbf{12.6} & \textbf{28.4} & 0 & 0 & 0 & 7.2 & 9.2 & 28 \\
 & Llama-2-7b+lora(BI) & 4.88 & 4.31 & 7.73 & 4.08 & 2 & 6.01 & 19.16 & 2.47 & 0.21 & 3.51 & 1.3 & 5.14 & 0.04 & 7.96 & 2.89 & 4.83 & 1.39 & 0.14 & 0.27 & 1.61 & 2.65 & 9.1 \\
 & Llama-2-7b+lora(Multi) & 5.22 & 5.7 & 3.77 & 5.11 & 3.19 & 6.36 & 15.84 & 2.57 & 0.27 & 4.07 & 1.72 & 6.31 & 0.1 & 8.17 & 3.62 & 5.56 & 1.06 & 0.03 & 0.53 & 1.4 & 2.8 & 11.32 \\
 & Llama-2-13b+lora(BI) & 9.16 & 8.29 & 9.97 & 9.93 & 3.81 & 10.25 & 21.06 & 3.33 & 0.61 & 6.94 & 2.29 & 8.61 & 1.05 & 11.37 & 4.81 & 9.1 & 0.16 & 0.31 & 0.28 & 2.95 & 5.22 & 17.49 \\
 & Llama-2-13b+lora(Multi) & 8.24 & 6.71 & 5.34 & 8.2 & 4.56 & 9.45 & 19.36 & 2.83 & 0.46 & 4.68 & 2.42 & 7.66 & 0.99 & 10.42 & 4.38 & 9.6 & 2.33 & 0.09 & 1.59 & 1.84 & 3.89 & 16.88 \\
 & Llama-2-13b+FF+lora(Multi) & 15.89 & 14.31 & \textbf{13.74} & 25.42 & 11.42 & 18.52 & 23.74 & \textbf{5.73} & \textbf{4.66} & 14.83 & 4.76 & 15.87 & \textbf{8.46} & \textbf{18.58} & 11.17 & 22.38 & \textbf{5.73} & \textbf{8.83} & \textbf{6.52} & 5.38 & 9.06 & 30.35 \\
 & Mistral-7B-v0.1+lora(Multi) & 5.25 & 6.46 & 2.03 & 4.18 & 2.64 & 6.06 & 15.63 & 2.14 & 0.11 & 2.77 & 2.12 & 6.11 & 0.02 & 6.92 & 1.54 & 5.68 & 1.05 & 0.01 & 0.54 & 1.43 & 1.75 & 8.71 \\ \midrule
 
\multirow{11}{*}{\textit{\textbf{IN22\_gen}}} & GPT-3.5 & 2.9 & 8.7 & 0.2 & 2.8 & 1.4 & 8.4 & 22.4 & 4.6 & 0.6 & 4.6 & 3.3 & 5.5 & - & 8 & 3.4 & 9.6 & 0.9 & - & 0.1 & 3.5 & 5.7 & 20 \\
 & IndicTrans-2 & \textbf{17.4} & \textbf{16.4} & 15.1 & \textbf{29.4} & \textbf{18.3} & \textbf{25.4} & \textbf{32.8} & \textbf{14.8} & 6.4 & \textbf{18.1} & \textbf{12.4} & \textbf{21.2} & 9.8 & 15.4 & 11.7 & \textbf{22.1} & 8.5 & 5.3 & \textbf{13.3} & \textbf{14} & \textbf{18.2} & \textbf{45.9} \\
 & Google Translate & 13.8 & - & - & 19.8 & 11.4 & 22.7 & 29.1 & 11.6 & - & 8.4 & 10.5 & 15.6 & - & 12.6 & 9.9 & - & - & - & - & \textbf{14} & 16.9 & 40.6 \\
 & LTRC, IIIT-H & - & - & - & - & - & 14 & 24.7 & 6 & - & - & 4.8 & 9.5 & - & - & - & - & - & - & - & 10 & 12.5 & 26.3 \\
 & SeamlessM4T & 12.6 & 13 & 0 & 0 & 0 & 19.4 & 27.4 & 11.3 & 0 & 14.4 & 10 & 14.7 & 0 & 14.1 & 13.6 & 21.6 & 0 & 2.3 & 0.5 & 13 & 15.7 & 35.3 \\
 & Llama-2-7b+lora(BI) & 6.22 & 5.84 & 8.48 & 5.06 & 3.1 & 5.19 & 20.16 & 4.41 & 0.63 & 4.51 & 2.95 & 8.21 & 0.2 & 7 & 4.57 & 4.23 & 3.54 & 0.14 & 1.01 & 2.99 & 3.86 & 10.68 \\
 & Llama-2-7b+lora(Multi) & 8.99 & 7.78 & 6.02 & 7.93 & 6.42 & 8 & 17.01 & 6.6 & 1.32 & 7.21 & 4.52 & 10.03 & 0.17 & 8.37 & 5.65 & 5.24 & 3.35 & 0.05 & 2.66 & 3.05 & 4.93 & 12.13 \\
 & Llama-2-13b+lora(BI) & 9.65 & 9.55 & 12 & 10.53 & 6.3 & 8.39 & 23.12 & 7.12 & 1.73 & 8.17 & 4.5 & 10.9 & 3.19 & 10.84 & 8.02 & 6.9 & 0.7 & 0.55 & 2.82 & 5.12 & 6.46 & 18.75 \\
 & Llama-2-13b+lora(Multi) & 10.66 & 9.7 & 7.46 & 10.98 & 8.88 & 9.73 & 20.66 & 7.45 & 1.97 & 7.12 & 5.7 & 12.34 & 2.02 & 10.46 & 7.56 & 7.67 & 4.93 & 0.08 & 4.66 & 4.44 & 6.03 & 17.1 \\
 & Llama-2-13b+FF+lora(Multi) & 17.18 & 16.11 & \textbf{16.08} & 27.4 & 15.06 & 16.26 & 27.01 & 14.22 & \textbf{7.1} & 17.53 & 11.3 & 20.31 & \textbf{11.72} & \textbf{17.39} & \textbf{15.2} & 15.74 & \textbf{10.4} & \textbf{7.07} & 11.3 & 10.75 & 12.55 & 32.88 \\
 & Mistral-7B-v0.1+lora(Multi) & 8.07 & 7.1 & 3.63 & 7.04 & 6.37 & 7.78 & 16.04 & 4.81 & 0.61 & 5.87 & 3.65 & 9.59 & 0.03 & 7.23 & 3.06 & 4.37 & 2.86 & 0.03 & 2.65 & 2.4 & 4.04 & 8.05 \\ \midrule

\multirow{11}{*}{\textit{\textbf{flores200-dev}}} & GPT-3.5 & 1.6 & 8.4 & - & - & - & 8.6 & 23.3 & 6.9 & 0.5 & 4.3 & 3.3 & 4 & 0 & 6.6 & 2.1 & 11.6 & 0.5 & 0 & 0 & 2.9 & 5.3 & 14.9 \\
 & IndicTrans-2 & \textbf{9.5} & \textbf{21} & - & - & - & \textbf{27.1} & \textbf{36.8} & 21 & \textbf{7.7} & \textbf{17.5} & 20.6 & 19.3 & - & \textbf{22.8} & 16 & \textbf{28.9} & \textbf{2.6} & 3.3 & 0 & \textbf{23} & 25.1 & 27.3 \\
 & Google Translate & 7.7 & - & - & - & - & 26.6 & \textbf{36.8} & \textbf{22.9} & - & 9.7 & \textbf{22.1} & \textbf{20.6} & - & 21.3 & \textbf{24.6} & - & - & - & - & 22.4 & \textbf{25.4} & \textbf{27.4} \\
 & LTRC, IIIT-H & - & - & - & - & - & 18.1 & 33.1 & 10 & - & - & 4.1 & 14.4 & - & - & - & - & - & - & - & 15.9 & 20.4 & 17.7 \\
 & SeamlessM4T & 9 & 18.5 & - & - & - & 24 & 35.3 & 19.8 & 0 & 14.4 & 16.6 & 18.1 & 0 & 18.5 & 17.2 & 27.8 & 0 & 0 & 0 & 20.3 & 23 & 24 \\
 & Llama-2-7b+lora(BI) & 2.8 & 4.88 & - & - & - & 5.34 & 22.78 & 2.94 & 0.29 & 2.89 & 1.97 & 4.69 & 0.07 & 4.91 & 2.41 & 4.31 & 0.62 & 0 & 0.06 & 3.15 & 4.13 & 7.77 \\
 & Llama-2-7b+lora(Multi) & 3.82 & 6.11 & - & - & - & 6.71 & 17.53 & 4.01 & 0.76 & 5.23 & 2.35 & 6.03 & 0.11 & 5.9 & 3.46 & 5.31 & 0.66 & 0.06 & 0.1 & 3.31 & 4.8 & 8.79 \\
 & Llama-2-13b+lora(BI) & 5 & 8.18 & - & - & - & 9.28 & 24.9 & 5.65 & 0.82 & 5.53 & 4.22 & 7.22 & 0.09 & 7.84 & 4.73 & 7.76 & 0.09 & 0.4 & 0.06 & 5.75 & 7.13 & 12.69 \\
 & Llama-2-13b+lora(Multi) & 4.99 & 7.76 & - & - & - & 8.6 & 20.67 & 5.02 & 1.21 & 6.55 & 3.05 & 7.82 & \textbf{0.12} & 8.49 & 4.99 & 7.84 & 0.92 & 0.08 & 0.13 & 4.58 & 6.16 & 11.79 \\
 & Llama-2-13b+FF+lora(Multi) & 9.08 & 13.7 & - & - & - & 17.48 & 29.16 & 12.32 & 3.35 & 11.88 & 11.36 & 14.65 & 0.05 & 15.84 & 13.24 & 20.79 & 2.09 & \textbf{4.43} & \textbf{0.14} & 13.28 & 15.97 & 21.64 \\
 & Mistral-7B-v0.1+lora(Multi) & 3.01 & 5.26 & - & - & - & 6.64 & 15.75 & 3.58 & 0.39 & 3.89 & 1.85 & 5.23 & 0.05 & 5.24 & 1.61 & 4.33 & 0.47 & 0.01 & 0.09 & 2.26 & 3.18 & 5.51 \\ \midrule

\multirow{11}{*}{\textit{\textbf{flores200-devtest}}} & GPT-3.5 & 1.8 & 8.2 & - & - & - & 9.6 & 23.9 & 7 & 0.3 & 4.1 & 3 & 5.4 & 0 & 7.5 & 3.3 & 11.2 & 0.7 & 0 & 0 & 3.3 & 5.5 & 16.6 \\
 & IndicTrans-2 & \textbf{9.6} & \textbf{21.2} & - & - & - & \textbf{27.4} & \textbf{36.6} & 22.7 & \textbf{6.8} & \textbf{17.2} & 20.3 & 19.6 & - & \textbf{23.1} & 15.7 & \textbf{26.1} & \textbf{3} & 3.4 & 0 & \textbf{22.4} & \textbf{26.7} & \textbf{26.3} \\
 & Google Translate & 8.1 & - & - & - & - & 27 & 36.2 & \textbf{24.1} & - & 10.3 & \textbf{21.2} & \textbf{20.3} & - & 21.5 & \textbf{23.4} & - & - & - & - & 21 & 26.5 & 25.2 \\
 & LTRC, IIIT-H & - & - & - & - & - & 18 & 32.7 & 11.6 & - & - & 3.9 & 14.7 & - & - & - & - & - & - & - & 15.3 & 20.9 & 17 \\
 & SeamlessM4T & 8.8 & 18.8 & - & - & - & 24.4 & 34.8 & 20.5 & 0 & 14.6 & 16.6 & 17.8 & 0 & 19.6 & 16.4 & 25.3 & 0 & 0 & 0 & 19.7 & 24.4 & 22.9 \\
 & Llama-2-7b+lora(BI) & 3 & 4.93 & - & - & - & 6.09 & 21.9 & 3.41 & 0.27 & 3.15 & 2.37 & 4.83 & 0.07 & 5.24 & 2.23 & 4.45 & 0.37 & 0.09 & 0.08 & 2.94 & 4.44 & 7.02 \\
 & Llama-2-7b+lora(Multi) & 3.63 & 5.92 & - & - & - & 6.89 & 16.77 & 4.2 & 0.62 & 5.22 & 2.58 & 5.91 & 0.17 & 6.25 & 3.47 & 5.11 & 0.51 & 0.03 & 0.14 & 2.92 & 5.24 & 7.78 \\
 & Llama-2-13b+lora(BI) & 4.69 & 8.11 & - & - & - & 9.31 & 23.71 & 5.97 & 0.9 & 5.41 & 4.08 & 7.28 & 0.14 & 8.94 & 4.47 & 7.24 & 0.08 & 0.35 & 0.15 & 5.58 & 7.4 & 12.31 \\
 & Llama-2-13b+lora(Multi) & 4.89 & 8.3 & - & - & - & 9.14 & 19.88 & 5.27 & 0.98 & 6.18 & 3.26 & 7.3 & \textbf{0.25} & 7.74 & 4.45 & 7.42 & 0.98 & 0.04 & 0.16 & 4.62 & 6.86 & 11.81 \\
 & Llama-2-13b+FF+lora(Multi) & 9.33 & 13.55 & - & - & - & 17.22 & 28.5 & 13.27 & 3.32 & 11.76 & 11.34 & 14.56 & 0.06 & 16.21 & 12.9 & 19.64 & 2.06 & \textbf{4.27} & \textbf{0.28} & 12.78 & 16.61 & 25.96 \\
 & Mistral-7B-v0.1+lora(Multi) & 3.26 & 5.04 & - & - & - & 6.39 & 15.07 & 3.49 & 0.31 & 3.78 & 1.96 & 5.06 & 0.08 & 5.69 & 1.53 & 4.59 & 0.53 & 0.01 & 0.11 & 2.37 & 3.55 & 5.19 \\ \midrule

\multirow{11}{*}{\textit{\textbf{newstest2019}}} & GPT-3.5 & - & 7.6 & - & - & - & 5.7 & 18.5 & 4.4 & - & - & 2 & 3.1 & - & - & - & 9 & - & - & - & 1.9 & 3.2 & - \\
 & IndicTrans-2 & - & \textbf{18.6} & - & - & - & \textbf{18.4} & 28 & 18.5 & - & - & 12 & 13.2 & - & - & - & \textbf{22.5} & - & - & - & 10.5 & 12.6 & - \\
 & Google Translate & - & - & - & - & - & \textbf{18.4} & \textbf{28.3} & \textbf{20.2} & - & - & \textbf{12.2} & \textbf{13.4} & - & - & - & - & - & - & - & \textbf{10.7} & \textbf{13.1} & - \\
 & LTRC, IIIT-H & - & - & - & - & - & 13.8 & 25.8 & 7.6 & - & - & 2.1 & 9 & - & - & - & - & - & - & - & 7.8 & 9.6 & - \\
 & SeamlessM4T & - & 17.6 & - & - & - & 18.2 & 27.6 & - & - & - & 9.7 & 12.7 & - & - & - & - & - & - & - & 9.9 & 11.6 & - \\
 & Llama-2-7b+lora(BI) & - & 3.64 & - & - & - & 3.54 & 16.15 & 1.93 & - & - & 1.07 & 2.76 & - & - & - & 3.58 & - & - & - & 1.44 & 2.2 & - \\
 & Llama-2-7b+lora(Multi) & - & 4.53 & - & - & - & 4.38 & 13.5 & 2.84 & - & - & 1.21 & 3.72 & - & - & - & 4.17 & - & - & - & 1.64 & 2.82 & - \\
 & Llama-2-13b+lora(BI) & - & 6.55 & - & - & - & 6.3 & 19.09 & 4.26 & - & - & 2.5 & 4.67 & - & - & - & 6.09 & - & - & -& 2.83 & 3.98 & - \\
 & Llama-2-13b+lora(Multi) & - & 6.26 & - & - & - & 6.31 & 16.34 & 3.43 & - & - & 2.1 & 4.99 & - & - & - & 6.07 & - & - & - & 2.48 & 3.61 & - \\
 & Llama-2-13b+FF+lora(Multi) & - & 12.52 & - & - & - & 12.25 & 22.28 & 9.69 & - & - & 11.34 & 9.39 & - & - & - & 16.32 & - & - & - & 6.62 & 8.25 & - \\
 & Mistral-7B-v0.1+lora(Multi) & - & 4.36 & - & - & - & 4.36 & 12.56 & 2.13 & - & - & 1.24 & 3.19 & - & - & - & 3.67 & - & - & - & 1.25 & 1.82 & - \\ \bottomrule
\end{tabular}%
}
\caption{BLEU scores across Models and Benchmark-sets; English to 22 Indian Languages; \\The symbol `-' indicates that the benchmark dataset for a particular language or machine translation system was not available during the evaluation period. Here, LORA stands for Low-Rank Adaptation of Large Language Models based fine-tuning. Multi stands for the multilingual model, FF for full-finetuning, and FF+lora stands for 2-stage fine-tuning.}
\label{tab:bleu_en_xx}
\end{table*}
% \end{landscape}
% Please add the following required packages to your document preamble:
% \usepackage{multirow}
% \usepackage{graphicx}
% \usepackage{lscape}
\begin{table*}[h]
\centering
\resizebox{\columnwidth}{!}{%
\begin{tabular}{llllllllllllllllllllllll}
\toprule
% \textbf{DataSet} & \multicolumn{1}{l|}{\textbf{Model}} & \multicolumn{1}{l|}{\textbf{Assamese}} & \multicolumn{1}{l|}{\textbf{Bangla}} & \multicolumn{1}{l|}{\textbf{Bodo}} & \multicolumn{1}{l|}{\textbf{Dogri}} & \multicolumn{1}{l|}{\textbf{Konkani}} & \multicolumn{1}{l|}{\textbf{Gujarati}} & \multicolumn{1}{l|}{\textbf{Hindi}} & \multicolumn{1}{l|}{\textbf{Kannada}} & \multicolumn{1}{l|}{\textbf{Kashmiri}} & \multicolumn{1}{l|}{\textbf{Maithili}} & \multicolumn{1}{l|}{\textbf{Malayalam}} & \multicolumn{1}{l|}{\textbf{Marathi}} & \multicolumn{1}{l|}{\textbf{Meitei}} & \multicolumn{1}{l|}{\textbf{Nepali}} & \multicolumn{1}{l|}{\textbf{Odia}} & \multicolumn{1}{l|}{\textbf{Punjabi}} & \multicolumn{1}{l|}{\textbf{Sanskrit}} & \multicolumn{1}{l|}{\textbf{Santali}} & \multicolumn{1}{l|}{\textbf{Sindhi}} & \multicolumn{1}{l|}{\textbf{Tamil}} & \multicolumn{1}{l|}{\textbf{Telugu}} & \textbf{Urdu} \\ \midrule

\textbf{DataSet} & \multicolumn{1}{l}{\textbf{Model}} & \multicolumn{1}{l}{\textbf{asm}} & \multicolumn{1}{l}{\textbf{ban}} & \multicolumn{1}{l}{\textbf{bod}} & \multicolumn{1}{l}{\textbf{doi}} & \multicolumn{1}{l}{\textbf{kon}} & \multicolumn{1}{l}{\textbf{guj}} & \multicolumn{1}{l}{\textbf{hin}} & \multicolumn{1}{l}{\textbf{kan}} & \multicolumn{1}{l}{\textbf{kas}} & \multicolumn{1}{l}{\textbf{mai}} & \multicolumn{1}{l}{\textbf{mal}} & \multicolumn{1}{l}{\textbf{mar}} & \multicolumn{1}{l}{\textbf{mei}} & \multicolumn{1}{l}{\textbf{nep}} & \multicolumn{1}{l}{\textbf{odi}} & \multicolumn{1}{l}{\textbf{pun}} & \multicolumn{1}{l}{\textbf{san}} & \multicolumn{1}{l}{\textbf{sat}} & \multicolumn{1}{l}{\textbf{sin}} & \multicolumn{1}{l}{\textbf{tam}} & \multicolumn{1}{l}{\textbf{tel}} & \textbf{urd} \\ \midrule

\multirow{11}{*}{\textit{\textbf{IN22\_conv}}} & GPT-3.5 & 25.4 & 41 & 0.1 & 11.7 & 8.3 & 37.3 & 46.1 & 29.3 & 6.3 & 24.2 & 29.8 & 32.8 & 0.3 & 42.3 & 26.1 & 40.1 & 19.2 & 0 & 0.1 & 32.1 & 34.3 & 48.3 \\
 & IndicTrans-2 & \textbf{48.8} & \textbf{49.8} & 47.7 & 51.2 & \textbf{45.1} & \textbf{54.7} & 49.8 & \textbf{36.5} & 28.2 & \textbf{46} & \textbf{44.1} & \textbf{51.4} & \textbf{42.4} & \textbf{54.3} & 41.8 & \textbf{56.2} & \textbf{38.9} & 37 & 29.9 & \textbf{43.3} & \textbf{48.6} & \textbf{60.1} \\
 & Google Translate & 45.2 & - & - & 39.5 & 43.3 & 53.5 & \textbf{50.9} & 36.1 & - & 37.6 & 43.7 & 49.4 & - & 49 & 40.2 & - & - & - & - & 43.1 & 47.9 & 59.4 \\
 & LTRC, IIIT-H & - & - & - & - & - & 44.5 & 45.4 & 31.2 & - & - & 33.8 & 42 & - & - & - & - & - & - & - & 38.6 & 42.5 & 48 \\
 & SeamlessM4T & 47.6 & 48.2 & 0 & 0 & 0 & 51.3 & 47.6 & 35 & 0 & 44.6 & 43.5 & 48.6 & 0 & 50.6 & \textbf{44.4} & 54.9 & 0 & 22 & 0.2 & 41.6 & 42.9 & 54.6 \\
 & Llama-2-7b+lora(BI) & 30.69 & 31.8 & 41.07 & 24.94 & 23.55 & 28.59 & 42.51 & 25.6 & 11.8 & 25.25 & 27.95 & 31.83 & 9.37 & 38.28 & 24.99 & 25.74 & 23.54 & 11.27 & 9.51 & 28 & 27.37 & 34.76 \\
 & Llama-2-7b+lora(Multi) & 31.9 & 34 & 32.16 & 28.28 & 27.98 & 26.8 & 39.54 & 26.61 & 11.93 & 28.6 & 29.03 & 33.7 & 4.51 & 39.24 & 27.79 & 24.6 & 22.89 & 0.68 & 10.18 & 28.15 & 29.21 & 36.6 \\
 & Llama-2-13b+lora(BI) & 37.33 & 38.53 & 45.06 & 34.33 & 29.87 & 35.95 & 44.58 & 28.25 & 19.45 & 31.7 & 33.39 & 38.17 & 19.46 & 43.8 & 31.33 & 32.65 & 11.87 & 12.6 & 10.13 & 34.75 & 34.56 & 43.9 \\
 & Llama-2-13b+lora(Multi) & 36.74 & 37.43 & 36.6 & 33.16 & 30.94 & 34.21 & 42.14 & 27.93 & 16.24 & 30.8 & 32.46 & 36.55 & 18.08 & 43.07 & 30.59 & 31.55 & 28.2 & 2.09 & 17.57 & 31.28 & 32.18 & 43.18 \\
 & Llama-2-13b+FF+lora(Multi) & 47.24 & 46.8 & \textbf{47.74} & \textbf{51.64} & 42.94 & 47 & 46.74 & 34.9 & \textbf{33.88} & 43.9 & 41.94 & 47.77 & 40.56 & 52.71 & 42.34 & 48.93 & 37.61 & \textbf{40.35} & \textbf{32.97} & 40.71 & 44.18 & 54.83 \\
 & Mistral-7B-v0.1+lora(Multi) & 31.65 & 35.3 & 26.27 & 26.17 & 27.07 & 28.91 & 39.19 & 24.45 & 8.71 & 26.33 & 29.9 & 33.78 & 3.54 & 38.5 & 12.33 & 24.31 & 22.95 & 0.11 & 11.88 & 28.07 & 25.71 & 34.07 \\ \midrule
 
\multirow{10}{*}{\textit{\textbf{IN22\_gen}}} & GPT-3.5 & 27.3 & 41.2 & 0.5 & 16.8 & 18.3 & 37.5 & 49.4 & 37.4 & 11.1 & 34.3 & 33.5 & 34.5 & 0.2 & 41.5 & 28.9 & 36.1 & 21.2 & 0.1 & 0.5 & 34 & 35.9 & 48.5 \\
 & IndicTrans-2 & \textbf{49.5} & \textbf{52.4} & \textbf{51.7} & \textbf{57.1} & \textbf{48.4} & \textbf{56.4} & \textbf{58} & \textbf{54.2} & 35.4 & \textbf{52} & \textbf{52.8} & \textbf{54.5} & \textbf{47.4} & \textbf{53.1} & 46.5 & \textbf{50.3} & 41.8 & \textbf{36.7} & \textbf{37.8} & \textbf{55.2} & \textbf{56.4} & \textbf{68.5} \\
 & Google Translate & 47.8 & - & - & 48.3 & 45.6 & 55.2 & 56.5 & 51.8 & - & 42.6 & 51.4 & 50.7 & - & 49.4 & 43.9 & - & - & - & - & 54.2 & 54.9 & 64.4 \\
 & LTRC, IIIT-H & - & - & - & - & - & 45.5 & 52.4 & 37.8 & - & - & 32.5 & 41.6 & - & - & - & - & - & - & - & 48.5 & 47.8 & 52.9 \\
 & SeamlessM4T & 46.3 & 49 & 0 & 0 & 0 & 52.4 & 55.2 & 51.2 & 0 & 48.9 & 49.4 & 49.3 & 0.3 & 50.6 & \textbf{48.3} & 49.7 & 0 & 18.1 & 0.8 & 53.2 & 53.1 & 63.2 \\
 & Llama-2-7b+lora(BI) & 29.4 & 33.06 & 40.57 & 27.15 & 25.68 & 28.7 & 46.26 & 30.45 & 15.09 & 28.54 & 27.99 & 34.93 & 9.25 & 36.27 & 26.41 & 24.71 & 26.65 & 11.23 & 14.2 & 31.77 & 29.67 & 38 \\
 & Llama-2-7b+lora(Multi) & 33.59 & 36.83 & 34.04 & 33.71 & 31.49 & 29.88 & 42.47 & 33.19 & 17.6 & 36.03 & 31.39 & 38.26 & 2 & 40.61 & 30.11 & 24.28 & 27.88 & 0.29 & 16.6 & 32 & 32.4 & 39.23 \\
 & Llama-2-13b+lora(BI) & 36.3 & 40.29 & 46.66 & 35.3 & 32.4 & 35.65 & 49.37 & 36.46 & 22.64 & 35.09 & 35.05 & 40.98 & 20.67 & 42.83 & 33.35 & 30.26 & 14.34 & 13.73 & 20.97 & 39.2 & 35.88 & 46.18 \\
 & Llama-2-13b+lora(Multi) & 36.9 & 40.47 & 37.07 & 37.91 & 34.61 & 34.69 & 46.34 & 36 & 21.2 & 37.95 & 35 & 41.41 & 15.03 & 44.21 & 33.03 & 29.67 & 32.06 & 1.05 & 22.45 & 36.8 & 35.15 & 44.47 \\
 & Llama-2-13b+FF+lora(Multi) & 46.87 & 49.57 & 50.63 & 54.64 & 45.17 & 46.1 & 53.61 & 47.45 & \textbf{37.18} & 50.59 & 48.19 & 52.18 & 41.85 & 52.87 & 44.47 & 42.54 & \textbf{42.04} & 36.44 & 36.05 & 50.2 & 46.68 & 58.3 \\
 & Mistral-7B-v0.1+lora(Multi) & 32.56 & 35.25 & 29.13 & 31.89 & 30.27 & 30.93 & 42.08 & 30.29 & 13.84 & 33.76 & 28.91 & 36.76 & 2.7 & 39.19 & 16.67 & 20.29 & 26.31 & 0.11 & 18.15 & 29.01 & 29.23 & 33.67 \\ \midrule
 
\multirow{11}{*}{\textit{\textbf{flores200-dev}}} & GPT-3.5 & 24.1 & 42.5 & - & - & - & 37.8 & 51 & 41.2 & 11.8 & 34.3 & 34.5 & 33.1 & - & 42.2 & 29.6 & 38.7 & 22 & 0.1 & - & 35.2 & 36.2 & 43 \\
 & IndicTrans-2 & \textbf{44.7} & \textbf{56.4} & - & - & - & \textbf{58} & 61.7 & 59.3 & \textbf{40.3} & \textbf{54.7} & \textbf{61.7} & 55 & - & \textbf{60.7} & 53.8 & \textbf{55.3} & \textbf{35.5} & 31.1 & - & \textbf{63.8} & 61.8 & \textbf{54} \\
 & Google Translate & 41.9 & - & - & - & - & 57.5 & \textbf{61.9} & \textbf{59.8} & - & 45 & 61.4 & \textbf{55.1} & - & 59 & \textbf{58.6} & - & - & - & - & 62.6 & \textbf{62} & 53.7 \\
 & LTRC, IIIT-H & - & - & - & - & - & 48.5 & 58.2 & 45.7 & - & - & 35.8 & 47.7 & - & - & - & - & - & - & - & 55.6 & 56.7 & 44.6 \\
 & SeamlessM4T & 42.6 & 53.7 & - & - & - & 55.3 & 60.3 & 57.4 & - & 50.5 & 57.4 & 53 & - & 56.7 & 54 & 54.2 & - & 22.9 & - & 60.9 & 59.1 & 51.9 \\
 & Llama-2-7b+lora(BI) & 28.67 & 33.84 & - & - & - & 30.91 & 48.94 & 33.87 & 13.19 & 26.58 & 30.05 & 34.64 & 1.2 & 37.17 & 28.48 & 25.7 & 22.18 & - & 0.25 & 34.21 & 32.45 & 34.35 \\
 & Llama-2-7b+lora(Multi) & 31.63 & 37.1 & - & - & - & 31.2 & 44.57 & 36.59 & 17.6 & 34.64 & 32.69 & 37.58 & 3.77 & 40.28 & 32.5 & 26.24 & 23.36 & 0 & \textbf{1.83} & 34.18 & 35.06 & 35.29 \\
 & Llama-2-13b+lora(BI) & 34.86 & 40.58 & - & - & - & 38.17 & 51.7 & 39.99 & 20.6 & 32.96 & 38.35 & 40.54 & 0.67 & 42.7 & 36.1 & 32.31 & 13.05 & 12.89 & 0.27 & 42.25 & 39.6 & 40.26 \\
 & Llama-2-13b+lora(Multi) & 35.03 & 40.51 & - & - & - & 36.84 & 48.24 & 39.8 & 20.79 & 37.78 & 36.91 & 40.89 & 3.7 & 44.56 & 36.14 & 31.15 & 26.88 & 1.1 & 0.74 & 39.6 & 38.27 & 38.96 \\
 & Llama-2-13b+FF+lora(Multi) & 42.89 & 49.4 & - & - & - & 48.5 & 55.61 & 51.66 & 32.15 & 48 & 52.69 & 50.1 & 0.29 & 54.63 & 50.1 & 47.7 & 33.88 & \textbf{32.59} & 0.44 & 55.6 & 53.11 & 49.26 \\
 & Mistral-7B-v0.1+lora(Multi) & 30.41 & 35.54 & - & - & - & 31.69 & 43.29 & 33.55 & 13.76 & 31.68 & 30.7 & 36.25 & \textbf{4.28} & 38.7 & 16.57 & 22.04 & 22.58 & 0.08 & 0.6 & 31.14 & 30.11 & 31 \\ \midrule
 
\multirow{11}{*}{\textit{\textbf{flores200-devtest}}} & GPT-3.5 & 24.4 & 41.4 & - & - & - & 39 & 51.1 & 41.5 & 9.7 & 33.7 & 33.9 & 35.8 & - & 43.2 & 30.8 & 38.2 & 23.4 & 0.1 & - & 36 & 36.8 & 44.2 \\
 & IndicTrans-2 & \textbf{44.5} & \textbf{56.1} & - & - & - & \textbf{59} & 60.8 & 60.1 & \textbf{39.5} & \textbf{54.5} & \textbf{61.9} & \textbf{55.1} & - & \textbf{60.6} & 53.1 & \textbf{53.2} & \textbf{36.1} & 30.9 & - & \textbf{62.8} & \textbf{63.1} & \textbf{53} \\
 & Google Translate & 42.1 & - & - & - & - & 58.4 & \textbf{61.1} & \textbf{60.5} & - & 45.3 & 61.7 & 54.9 & - & 59.2 & \textbf{57.6} & - & - & - & - & 61.5 & 62.6 & 51.9 \\
 & LTRC, IIIT-H & - & - & - & - & - & 49.4 & 57.8 & 46.2 & - & - & 36.1 & 48.2 & - & - & - & - & - & - & - & 55.1 & 57.4 & 44.1 \\
 & SeamlessM4T & 42.4 & 54 & - & - & - & 56.3 & 59.5 & 58.2 & - & 50.6 & 57.6 & 52.5 & - & 56.4 & 53.4 & 52.2 & 0 & 22.5 & 0 & 60.3 & 60.4 & 51.1 \\
 & Llama-2-7b+lora(BI) & 28.11 & 33.03 & - & - & - & 30.63 & 48.1 & 33.24 & 12.76 & 26.03 & 29.4 & 34.87 & 1.2 & 36.17 & 27.63 & 25.26 & 21.9 & 11.5 & 0.28 & 34 & 31.94 & 33.51 \\
 & Llama-2-7b+lora(Multi) & 31.57 & 36.3 & - & - & - & 30.85 & 43.18 & 36.6 & 17.35 & 33.96 & 32.15 & 38.04 & 4.28 & 40.18 & 31.78 & 25.2 & 23.95 & 0 & \textbf{2.17} & 34.25 & 35.08 & 33.88 \\
 & Llama-2-13b+lora(BI) & 33.95 & 39.61 & - & - & - & 37.96 & 50.26 & 39.55 & 19.93 & 31.87 & 37.03 & 40.25 & 0.48 & 42.64 & 35.4 & 30.93 & 12.56 & 12.79 & 0.37 & 42.11 & 39.28 & 39.85 \\
 & Llama-2-13b+lora(Multi) & 35.09 & 39.93 & - & - & - & 36.8 & 47.24 & 39.35 & 19.79 & 37.03 & 35.94 & 40.49 & 3.74 & 43.99 & 35.11 & 29.35 & 27.44 & 0.85 & 0.87 & 39.13 & 38.21 & 38.9 \\
 & Llama-2-13b+FF+lora(Multi) & 43.3 & 48.51 & - & - & - & 48.76 & 54.89 & 51.53 & 32.23 & 47.83 & 53.08 & 50.18 & 0.24 & 55.01 & 49.78 & 46.69 & 34.86 & \textbf{32.4} & 0.51 & 54.54 & 53.36 & 48.4 \\
 & Mistral-7B-v0.1+lora(Multi) & 30.63 & 34.08 & - & - & - & 31.2 & 41.78 & 33 & 13.56 & 31.14 & 30.16 & 35.43 & \textbf{4.33} & 38.73 & 17.84 & 23.31 & 22.7 & 0.07 & 0.6 & 31.34 & 29.83 & 29.67 \\ \midrule

\multirow{11}{*}{\textit{\textbf{newstest2019}}} & GPT-3.5 & - & 41.7 & - & - & - & 35.5 & 45.2 & 38 & - & - & 31.5 & 32.1 & - & - & - & 35.8 & - & - & - & 31.3 & 33.4 & 16.2 \\
 & IndicTrans-2 & - & \textbf{55.7} & - & - & - & \textbf{52.4} & 54.1 & 57.3 & - & - & \textbf{53} & \textbf{51} & - & - & - & 51.1 & - & - & - & \textbf{51.3} & 49.7 & \textbf{17.6} \\
 & Google Translate & - & - & - & - & - & 51.7 & \textbf{54.5} & \textbf{57.4} & - & - & 52.2 & 49.8 & - & - & - & - & - & - & - & 50.4 & \textbf{50.2} & 17.4 \\
 & LTRC, IIIT-H & - & - & - & - & - & 46 & 51.9 & 39.2 & - & - & 31.8 & 42.9 & - & - & - & - & - & - & - & 45.8 & 44.3 & 16.5 \\
 & SeamlessM4T & - & 54 & - & - & - & 51.3 & 53.3 & 55.7 & - & - & 49.7 & 49.1 & - & - & - & \textbf{51.5} & - & - & - & 49.9 & 48.1 & 17.5 \\
 & Llama-2-7b+lora(BI) & - & 31.65 & - & - & - & 27.26 & 42.24 & 30.16 & - & - & 26.57 & 31.24 & - & - & - & 23.25 & - & - & - & 28.76 & 27.69 & - \\
 & Llama-2-7b+lora(Multi) & - & 34.85 & - & - & - & 28.34 & 39.63 & 33.88 & - & - & 28.4 & 34.95 & - & - & - & 23.61 & - & - & - & 28.7 & 30.5 & - \\
 & Llama-2-13b+lora(BI) & - & 38.9 & - & - & - & 34.34 & 45.35 & 36.54 & - & - & 32.99 & 37.39 & - & - & - & 28.61 & - & - & - & 35.4 & 33.9 & - \\
 & Llama-2-13b+lora(Multi) & - & 38.96 & - & - & - & 33.77 & 43.21 & 36.25 & - & - & 32.51 & 37.68 & - & - & - & 27.11 & - & - & - & 33.83 & 32.9 & - \\
 & Llama-2-13b+FF+lora(Multi) & - & 48 & - & - & - & 43.68 & 48.19 & 47.1 & - & - & 45.1 & 45.51 & - & - & - & 42.83 & - & - & - & 45.58 & 42.54 & - \\
 & Mistral-7B-v0.1+lora(Multi) & - & 34.18 & - & - & - & 29.39 & 38.74 & 29.77 & - & - & 27.48 & 33.19 & - & - & - & 20.77 & - & - & - & 26.54 & 25.87 & - \\ \bottomrule
\end{tabular}%
}
\caption{chrF scores across Models and Benchmark-sets; English to 22 Indian Languages;\\The symbol `-' indicates that the benchmark dataset for a particular language or machine translation system was not available during the evaluation period.  Here, LORA stands for Low-Rank Adaptation of Large Language Models based fine-tuning. Multi stands for the multilingual model, FF for full-finetuning, and FF+lora stands for 2-stage fine-tuning.}
\label{tab:chrf_en_xx}
\end{table*}
% Please add the following required packages to your document preamble:
% \usepackage{multirow}
% \usepackage{graphicx}
% \usepackage{lscape}
\begin{table*}[h]
\centering
\resizebox{\columnwidth}{!}{%
\begin{tabular}{llllllllllllllllllllllll}
\toprule
% \textbf{DataSet} & \textbf{Model} & \textbf{Assamese} & \textbf{Bangla} & \textbf{Bodo} & \textbf{Dogri} & \textbf{Konkani} & \textbf{Gujarati} & \textbf{Hindi} & \textbf{Kannada} & \textbf{Kashmiri} & \textbf{Maithili} & \textbf{Malayalam} & \textbf{Marathi} & \textbf{Meitei} & \textbf{Nepali} & \textbf{Odia} & \textbf{Punjabi} & \textbf{Sanskrit} & \textbf{Santali} & \textbf{Sindhi} & \textbf{Tamil} & \textbf{Telugu} & \textbf{Urdu} \\ \hline

\textbf{DataSet} & \multicolumn{1}{l}{\textbf{Model}} & \multicolumn{1}{l}{\textbf{asm}} & \multicolumn{1}{l}{\textbf{ban}} & \multicolumn{1}{l}{\textbf{bod}} & \multicolumn{1}{l}{\textbf{doi}} & \multicolumn{1}{l}{\textbf{kon}} & \multicolumn{1}{l}{\textbf{guj}} & \multicolumn{1}{l}{\textbf{hin}} & \multicolumn{1}{l}{\textbf{kan}} & \multicolumn{1}{l}{\textbf{kas}} & \multicolumn{1}{l}{\textbf{mai}} & \multicolumn{1}{l}{\textbf{mal}} & \multicolumn{1}{l}{\textbf{mar}} & \multicolumn{1}{l}{\textbf{mei}} & \multicolumn{1}{l}{\textbf{nep}} & \multicolumn{1}{l}{\textbf{odi}} & \multicolumn{1}{l}{\textbf{pun}} & \multicolumn{1}{l}{\textbf{san}} & \multicolumn{1}{l}{\textbf{sat}} & \multicolumn{1}{l}{\textbf{sin}} & \multicolumn{1}{l}{\textbf{tam}} & \multicolumn{1}{l}{\textbf{tel}} & \textbf{urd} \\ \midrule

\multirow{11}{*}{\textit{\textbf{IN22\_conv}}} & GPT-3.5 & 19.9 & 29.8 & 2.8 & 19.9 & 9.7 & 28.3 & 34.1 & 17.8 & 6.2 & 14 & 20.7 & 24 & 0.4 & 29.3 & 21.1 & 30.9 & 14.5 & 0.2 & 9.6 & 15.4 & 19.9 & 34.7 \\
& IndicTrans-2 & 43.9 & 36.9 & \textbf{35.3} & \textbf{45.5} & 28.9 & \textbf{40.9} & 38.7 & \textbf{25.1} & \textbf{31.8} & 35.3 & \textbf{31.3} & 37 & \textbf{32.5} & \textbf{43.1} & \textbf{38.9} & \textbf{42.8} & 25.8 & \textbf{24.2} & \textbf{26} & 22.6 & 30.8 & \textbf{46.1} \\
& Google Translate & \textbf{44.5} & \textbf{37.6} & 1.8 & 42.3 & \textbf{29.5} & \textbf{40.9} & \textbf{39.4} & 24.3 & 5.6 & \textbf{36.4} & 31.1 & \textbf{37.6} & 25.7 & 43 & 37.4 & 39.4 & \textbf{26.7} & 0 & 8.7 & \textbf{23.3} & \textbf{31.5} & 45.6 \\
& LTRC, IIIT-H & - & - & - & - & - & - & 23.9 & 10.5 & - & - & 14.6 & 19.1 & - & - & - & - & - & - & - & - & 14.1 & - \\
& SeamlessM4T & 41 & 35.9 & 0 & 0 & 0 & 40.6 & 38.1 & 23 & 0 & 34 & 30.3 & 35.5 & 0.2 & 40.3 & 38.6 & 41.4 & 0 & 17.2 & 9.3 & 23.1 & 31.1 & 42.3 \\
& Llama-2-7b+lora(BI) & 1.17 & 2.42 & 4.04 & 9.49 & 4.43 & 3.88 & 15.72 & 1.45 & 2.17 & 3.08 & 1.7 & 7 & 0.08 & 6.22 & 1.07 & 2.29 & 4.62 & 0.03 & 3.45 & 0.81 & 0.84 & 6.33 \\
& Llama-2-7b+lora(Multi) & 12.43 & 7.71 & 8.9 & 10.16 & 6.14 & 5.09 & 7.68 & 4.25 & 4.33 & 9.79 & 3.34 & 9.91 & 1.15 & 12.36 & 6.23 & 6.55 & 5.14 & 0.26 & 3.75 & 3.47 & 5.33 & 13.75 \\
& Llama-2-13b+lora(BI) & 2.49 & 3.12 & 15.01 & 0.9 & 2.11 & 1.26 & 25.04 & 0.82 & 2.93 & 2.86 & 8.71 & 9.44 & 0.31 & 1.49 & 1.07 & 1.61 & - & - & - & - & 1.34 & 6.72 \\
& Llama-2-13b+lora(Multi) & 21.19 & 20.26 & 15.51 & 18.4 & 13.37 & 17.06 & 23.66 & 8.15 & 8.54 & 15.77 & 12.85 & 17.37 & 2.93 & 22.89 & 11.77 & 15.41 & 9.85 & 0.72 & 8.17 & 9.27 & 11.53 & 24.16 \\
& Llama-2-13b+FF+lora(Multi) & 2.26 & 1.77 & 1.48 & 2 & 1.58 & 0.84 & 7.05 & 0.8 & 0.94 & 1.9 & 5.99 & 2.06 & 0.48 & 2.71 & 1.55 & 1.52 & 1.35 & 0.13 & 1.81 & 0.84 & 0.92 & 2.31 \\
& Mistral-7B-v0.1+lora(Multi) & 12.23 & 9.19 & 8.55 & 11.63 & 7.46 & 2.38 & 14.81 & 3.62 & 4.69 & 12.91 & 3.5 & 11.2 & 0.43 & 17.9 & 5.51 & 1.69 & 11.03 & 0.1 & 3.84 & 3.04 & 2.4 & 18.58 \\  \midrule
 
\multirow{11}{*}{\textit{\textbf{IN22\_gen}}} & GPT-3.5 & 19 & 25.2 & 6 & 20.8 & 13.4 & 25.6 & 30.4 & 23.6 & 10 & 19.5 & 15.8 & 22.5 & 0.2 & 27.6 & 18.9 & 25.6 & 14.3 & 0.2 & 13.7 & 14.4 & 20.2 & 29.2 \\
& IndicTrans-2 & \textbf{42.5} & \textbf{40.8} & \textbf{37.5} & \textbf{53.4} & \textbf{32.7} & \textbf{43.1} & \textbf{40} & 40 & \textbf{38.4} & \textbf{42.5} & \textbf{40.4} & \textbf{41.5} & \textbf{38.4} & \textbf{47.8} & \textbf{43.3} & \textbf{40.8} & \textbf{30.6} & \textbf{25} & \textbf{31.5} & \textbf{35.9} & \textbf{42.3} & \textbf{53.7} \\
& Google Translate & 41.9 & 39.9 & 4.3 & 44.8 & 33 & 43 & 39.2 & \textbf{41.1} & 9.7 & 39.7 & 37.9 & 40.6 & 27.4 & 46.8 & 40.3 & 39.6 & 28.5 & 0.2 & 15.7 & 34.8 & 40.9 & 51.3 \\
& LTRC, IIIT-H & - & - & - & - & - & - & 20.7 & 15.4 & - & - & 13.7 & 15.4 & - & - & - & - & - & - & - & - & 15.1 & - \\
& SeamlessM4T & 40.7 & 37.3 & 0 & 0 & 0 & 41.6 & 37.3 & 39.3 & 0 & 39.6 & 36.9 & 37.3 & 0 & 43.5 & 40.8 & 38.5 & 0 & 15.7 & 15 & 33.1 & 39.2 & 48.3 \\
& Llama-2-7b+lora(BI) & 2.53 & 5.99 & 8.3 & 11.99 & 7.13 & 6.13 & 18.21 & 3.62 & 4.95 & 7.41 & 5.28 & 10.72 & 0.16 & 8.81 & 2 & 4.51 & 6.82 & 0.09 & 6.97 & 4.11 & 3.18 & 12.37 \\
& Llama-2-7b+lora(Multi) & 12.01 & 9.89 & 8.79 & 14.42 & 9.61 & 5.07 & 11.38 & 8.51 & 7.64 & 14.69 & 6.39 & 11.47 & 0.77 & 16.75 & 7.71 & 6.14 & 9.36 & 0.4 & 8.1 & 5.82 & 7.36 & 16.98 \\
& Llama-2-13b+lora(BI) & 6.66 & 8.01 & 12.15 & 3.95 & 7.74 & 5.8 & 26.55 & 2.2 & 8.17 & 8.4 & 10.06 & 13.82 & 0.88 & 2.7 & 4.85 & 4.61 & - & - & - & - & 4.02 & 17.46 \\
& Llama-2-13b+lora(Multi) & 20.94 & 20.33 & 16.2 & 24.11 & 17.95 & 15.31 & 26.31 & 14.93 & 14.16 & 23.39 & 13.61 & 22.95 & 2.88 & 26.64 & 14.29 & 11.5 & 14.74 & 0.87 & 13.45 & 12.25 & 13.12 & 26.43 \\
& Llama-2-13b+FF+lora(Multi) & 1.33 & 1.78 & 1.56 & 2.17 & 1.74 & 1 & 3.61 & 0.88 & 1.12 & 1.83 & 2.56 & 2.12 & 0.22 & 2.06 & 0.89 & 1.02 & 1.06 & 0.21 & 2.41 & 1.39 & 1.04 & 1.68 \\
& Mistral-7B-v0.1+lora(Multi) & 12.36 & 10.9 & 9.31 & 16.72 & 10.94 & 2.68 & 13.88 & 7.82 & 8.38 & 16.02 & 4.49 & 14.17 & 0.69 & 16.96 & 5.5 & 3.39 & 12.33 & 0.12 & 8.19 & 6.52 & 6.17 & 20.46 \\  \midrule

\multirow{11}{*}{\textit{\textbf{flores200-dev}}} & GPT-3.5 & 14.8 & 25.7 & - & - & - & 24.4 & 32.8 & 22.1 & 7.9 & 19.6 & 19.3 & 21.5 & 0 & 26.2 & 18.1 & 26.5 & 11.3 & 0.4 & 0 & 15.1 & 20.3 & 27.3 \\
 & IndicTrans-2 & \textbf{34.8} & 40.4 & - & - & - & 44.5 & 46.9 & 39.1 & \textbf{39} & \textbf{49} & 41.4 & 41.8 & - & 46.7 & 43.2 & 48.1 & \textbf{27} & \textbf{19.7} & - & 38.9 & 45.9 & 40 \\
 & Google Translate & 34 & \textbf{40.8} & - & - & - & \textbf{45.6} & \textbf{47.7} & \textbf{39.7} & 11.5 & 48.7 & \textbf{42} & \textbf{42.5} & - & \textbf{47.4} & \textbf{43.9} & \textbf{48.3} & 25.1 & 0.2 & - & \textbf{39.3} & \textbf{46.4} & \textbf{41.8} \\
 & LTRC, IIIT-H & - & - & - & - & - & - & 27.1 & 17.3 & - & - & 18.5 & 19.1 & - & - & - & - & - & - & - & - & 20.1 & - \\
 & SeamlessM4T & 34.2 & 39.2 & - & - & - & 0 & 0 & 37.7 & 0 & 46.3 & 40 & 39.7 & 0 & 44.5 & 41.2 & 45.8 & 0 & \textbf{19.7} & 0 & 37.5 & 43.6 & 39.8 \\
 & Llama-2-7b+lora(BI) & 1.88 & 5.57 & - & - & - & 5.57 & 17.98 & 3.37 & 4.82 & 6.97 & 5.77 & 10.82 & 1.36 & 8.41 & 2.24 & 5.61 & 6.73 & 0.17 & 2.32 & 2.97 & 2.87 & 9.35 \\
 & Llama-2-7b+lora(Multi) & 11.7 & 10.84 & - & - & - & 4.65 & 8.16 & 8.67 & 6.92 & 14.84 & 7.84 & 11.67 & 3.23 & 14.63 & 7.78 & 6.61 & 6.67 & 0.48 & 1.45 & 6.13 & 8.34 & 13 \\
 & Llama-2-13b+lora(BI) & 4.83 & 9.12 & - & - & - & 4.74 & 30.09 & 2.53 & 7.01 & 8.56 & 10.55 & 14.6 & 2.64 & 2.38 & 4.32 & 3.35 & - & - & - & 4 & 9.3 & - \\
 & Llama-2-13b+lora(Multi) & 18.4 & 22.13 & - & - & - & 15.47 & 29.09 & 14.67 & 12.48 & 24.99 & 15.75 & 22.2 & \textbf{5.65} & 24.27 & 14.71 & 17.31 & 12.36 & 0.94 & 4.51 & 14.7 & 15.39 & 22.23 \\
 & Llama-2-13b+FF+lora(Multi) & 1.34 & - & - & - & - & 0.87 & 3.35 & 1.31 & 0.8 & 1.71 & 1.78 & 2.2 & 1.15 & 1.71 & 1.07 & 0.86 & 1.17 & 0.18 & 0.84 & 1.29 & 1.31 & - \\
 & Mistral-7B-v0.1+lora(Multi) & 11.31 & 11.4 & - & - & - & 3.24 & 13.02 & 7.79 & 7.1 & 17.98 & 6 & 15.07 & 3.63 & 17.88 & 5.23 & 2.7 & 9.56 & 0.22 & 3.09 & 6.31 & 5.76 & 15.35 \\ \midrule
 
\multirow{11}{*}{\textit{\textbf{flores200-devtest}}} & GPT-3.5 & 14.5 & 24.1 & - & - & - & 23.6 & 32.5 & 20.8 & 8.2 & 18 & 18.5 & 21.6 & 0 & 24.8 & 16.2 & 24.4 & 11.1 & 0.4 & - & 14.2 & 16.7 & 25.4 \\
 & IndicTrans-2 & \textbf{33.1} & 39.3 & - & - & - & 45.2 & \textbf{46.1} & 37.7 & \textbf{36.2} & \textbf{48.3} & \textbf{41} & 41.5 & - & \textbf{46.3} & \textbf{42.6} & 44.7 & \textbf{26.8} & 18.1 & - & \textbf{37.8} & 44.8 & 38.1 \\
 & Google Translate & 32.8 & \textbf{39.8} & - & - & - & \textbf{46.2} & \textbf{46.1} & \textbf{38} & 10.7 & 46.6 & 40.9 & \textbf{42.1} & - & \textbf{46.3} & 41.3 & \textbf{45.9} & 25.2 & 0.1 & - & 37.7 & \textbf{44.9} & \textbf{40.1} \\
 & LTRC, IIIT-H & - & - & - & - & - & - & 27.6 & 16.7 & - & - & 17.9 & 18.2 & - & - & - & - & - & - & - & - & 19.2 & - \\
 & SeamlessM4T & 32.3 & 38.3 & - & - & - & - & - & 36 & - & 44.3 & 39.7 & 38.8 & - & 43.4 & 40.2 & 43.2 & 0 & \textbf{18.3} & 0 & 35.2 & 42.9 & 38.1 \\
 & Llama-2-7b+lora(BI) & 1.64 & 5.51 & - & - & - & 5.78 & 16.88 & 2.91 & 4.28 & 7.45 & 5.76 & 10.24 & 1.18 & 8.56 & 2.11 & 5.48 & 6.75 & 0.1 & 2.22 & 2.46 & 3.02 & 8.37 \\
 & Llama-2-7b+lora(Multi) & 10.85 & 10.45 & - & - & - & 4.55 & 8.1 & 8.34 & 6.76 & 14.02 & 6.39 & 11.26 & 2.66 & 15.03 & 7.02 & 5.9 & 6.82 & 0.55 & 1.66 & 5.94 & 7.53 & 13.94 \\
 & Llama-2-13b+lora(BI) & 4.29 & 8.76 & - & - & - & 4.15 & 30.38 & 2.57 & 5.94 & 7.49 & 10.07 & 14.12 & 2.12 & 2.51 & 3.36 & 3.48 & - & - & - & 3.66 & 8.98 & - \\
 & Llama-2-13b+lora(Multi) & 17.64 & 20.59 & - & - & - & 15.49 & 28.6 & 13.86 & 11.17 & 23.6 & 14.97 & 21.84 & \textbf{5.38} & 23.58 & 13.13 & 15.22 & 11.85 & 0.94 & \textbf{4.51} & 12.73 & 15.29 & 21.33 \\
 & Llama-2-13b+FF+lora(Multi) & 1.12 & 1.42 & - & - & - & 0.75 & 2.67 & 0.95 & 0.88 & 1.53 & 1.7 & 1.73 & 0.92 & 1.38 & 0.9 & 0.97 & 0.81 & 0.15 & 0.78 & 1.26 & 1.16 & 1.27 \\
 & Mistral-7B-v0.1+lora(Multi) & 10.21 & 10.86 & - & - & - & 3.22 & 12.62 & 7.3 & 7.29 & 15.45 & 5.2 & 14.16 & 3.71 & 15.91 & 5.8 & 2.66 & 11.08 & 0.33 & 3.31 & 5.82 & 6.15 & 13.87 \\ \midrule
 
\multirow{11}{*}{\textit{\textbf{newstest2019}}} & GPT-3.5 & - & 20 & - & - & - & 19.2 & 24.1 & 16 & - & - & 11.6 & 14.8 & - & - & - & 18.7 & - & - & - & 9.9 & 12.6 & 2.3 \\
 & IndicTrans-2 & - & \textbf{38.8} & - & - & - & 42.5 & \textbf{37.7} & 36.5 & - & - & 32.1 & 37 & - & - & - & \textbf{40.5} & - & - & - & 31.3 & 29.9 & 3.1 \\
 & Google Translate & - & 38.5 & - & - & - & \textbf{42.6} & - & \textbf{36.6} & - & - & \textbf{34} & \textbf{37.7} & - & - & - & - & - & - & - & \textbf{31.8} & \textbf{30.5} & \textbf{4.3} \\
 & LTRC, IIIT-H & - & - & - & - & - & - & 12.7 & 13.4 & - & - & 13.8 & 15.7 & - & - & - & - & - & - & - & - & 9.4 & - \\
 & SeamlessM4T & - & 35.3 & - & - & - & 39.6 & 35.3 & 33.8 & - & - & 30 & 33.4 & - & - & - & 37.6 & - & - & - & 29.4 & 27.9 & 3.8 \\
 & Llama-2-7b+lora(BI) & - & 5.14 & - & - & - & 4.14 & 12.5 & 2.51 & - & - & 4.55 & 8.97 & - & - & - & 4.12 & - & - & - & 2.32 & 2.57 & 0.86 \\
 & Llama-2-7b+lora(Multi) & - & 8.57 & - & - & - & 3.92 & 11.9 & 5.32 & - & - & 4.77 & 10.57 & - & - & - & 5.08 & - & - & - & 4.2 & 5.07 & 1.25 \\
 & Llama-2-13b+lora(BI) & - & 10.78 & - & - & - & 5.29 & 23.88 & 1.86 & - & - & 9.08 & 12.95 & - & - & - & 2.7 & - & - & - & - & 2.48 & 1.31 \\
 & Llama-2-13b+lora(Multi) & - & 18.27 & - & - & - & 12.4 & 24.46 & 10.42 & - & - & 9.91 & 18.52 & - & - & - & 10.79 & - & - & - & 8.91 & 9.83 & 2.06 \\
 & Llama-2-13b+FF+lora(Multi) & - & 1.15 & - & - & - & 0.69 & 2.42 & 0.9 & - & - & 1.29 & 1.46 & - & - & - & 0.88 & - & - & - & 0.99 & 0.88 & 0.16 \\
 & Mistral-7B-v0.1+lora(Multi) & - & 9.84 & - & - & - & 2.8 & 11.84 & 6.19 & - & - & 3.77 & 11.38 & - & - & - & 1.74 & - & - & - & 4.3 & 4.21 & 1.61 \\ \bottomrule 
\end{tabular}%
}
\caption{BLEU scores across Models and Benchmark-sets; 22 Indian Languages to English;\\The symbol `-' indicates that the benchmark dataset for a particular language or machine translation system was not available during the evaluation period.  Here, LORA stands for Low-Rank Adaptation of Large Language Models based fine-tuning. Multi stands for the multilingual model, FF for full-finetuning, and FF+lora stands for 2-stage fine-tuning.}
\label{tab:xx_en_bleu}
\end{table*}
% Please add the following required packages to your document preamble:
% \usepackage{multirow}
% \usepackage{graphicx}
% \usepackage{lscape}
\begin{table*}[h]
\centering
\resizebox{\columnwidth}{!}{%
\begin{tabular}{llllllllllllllllllllllll}
\toprule
% \textbf{DataSet} & \textbf{Model} & \textbf{Assamese} & \textbf{Bangla} & \textbf{Bodo} & \textbf{Dogri} & \textbf{Konkani} & \textbf{Gujarati} & \textbf{Hindi} & \textbf{Kannada} & \textbf{Kashmiri} & \textbf{Maithili} & \textbf{Malayalam} & \textbf{Marathi} & \textbf{Meitei} & \textbf{Nepali} & \textbf{Odia} & \textbf{Punjabi} & \textbf{Sanskrit} & \textbf{Santali} & \textbf{Sindhi} & \textbf{Tamil} & \textbf{Telugu} & \textbf{Urdu} \\ \hline
\textbf{DataSet} & \multicolumn{1}{l}{\textbf{Model}} & \multicolumn{1}{l}{\textbf{asm}} & \multicolumn{1}{l}{\textbf{ban}} & \multicolumn{1}{l}{\textbf{bod}} & \multicolumn{1}{l}{\textbf{doi}} & \multicolumn{1}{l}{\textbf{kon}} & \multicolumn{1}{l}{\textbf{guj}} & \multicolumn{1}{l}{\textbf{hin}} & \multicolumn{1}{l}{\textbf{kan}} & \multicolumn{1}{l}{\textbf{kas}} & \multicolumn{1}{l}{\textbf{mai}} & \multicolumn{1}{l}{\textbf{mal}} & \multicolumn{1}{l}{\textbf{mar}} & \multicolumn{1}{l}{\textbf{mei}} & \multicolumn{1}{l}{\textbf{nep}} & \multicolumn{1}{l}{\textbf{odi}} & \multicolumn{1}{l}{\textbf{pun}} & \multicolumn{1}{l}{\textbf{san}} & \multicolumn{1}{l}{\textbf{sat}} & \multicolumn{1}{l}{\textbf{sin}} & \multicolumn{1}{l}{\textbf{tam}} & \multicolumn{1}{l}{\textbf{tel}} & \textbf{urd} \\ \midrule

\multirow{11}{*}{\textit{\textbf{IN22\_conv}}} & GPT-3.5 & 44.80     & 54.30   & 21.10  & 45.50  & 32.80    & 52.60     & 58.80  & 45.10   & 28.70     & 42.10     & 46.00        & 50.30    & 14.50   & 53.80  & 47.60  & 54.80    & 40.60     & 14.80    & 34.70   & 39.90  & 44.10   & 59.00 \\
 & IndicTrans-2 & 63.90     & 59.80   & \textbf{57.20}  & \textbf{66.00}    & 52.60    & \textbf{63.20}     & 60.90  & 49.20   & \textbf{53.50}     & 59.20     & 55.40      & 60.10    & \textbf{53.90}   & 64.40   & \textbf{61.60}  & \textbf{63.40}    & 49.60     & \textbf{45.10}    & \textbf{50.60}   & 47.40  & 54.20   & 66.60 \\
 & Google Translate & \textbf{64.70}     & \textbf{60.80}   & 16.40  & 63.70  & \textbf{52.80}    & \textbf{63.20}    & \textbf{61.80}  & \textbf{49.70}   & 23.20     & \textbf{60.00}       & \textbf{56.10}      & \textbf{60.50 }   & 47.20   & \textbf{64.90}   & 60.1  & 62.70    & \textbf{50.30}     & 0.30     & 32.80   & \textbf{48.50}  & \textbf{55.20}   & \textbf{66.30} \\
 & LTRC, IIIT-H & -        & -      & -     & -     & -       & -        & 52.20  & 34.80   & -        & -        & 40.70      & 47.30    & -      & -      & -     & -       & -        & -       & -      & -     & 39.00     & -  \\
 & SeamlessM4T & 61.30     & 59.00     & -     & -     & -       & 62.70     & 60.50  & 47.90   & -        & 57.60     & 54.90      & 58.70    & 2.90    & 62.10   & 61.10  & 62.60    & -        & 37.20    & 32.00     & 47.80  & 53.90   & 63.90 \\
 & Llama-2-7b+lora(BI) & 1.49     & 4.57   & 8.27  & 24.13 & 16.23   & 9.36     & 31.77 & 4.18   & 15.60     & 7.80      & 4.99      & 15.83   & 1.20    & 12.66  & 2.94  & 5.45    & 13.11    & 0.29    & 20.31  & 2.24  & 2.45   & 16.30 \\
 & Llama-2-7b+lora(Multi) & 26.69    & 17.30   & 21.90  & 23.50  & 16.78   & 10.84    & 14.04 & 14.27  & 16.95    & 23.63    & 11.05     & 21.46   & 8.80    & 24.88  & 16.87 & 12.50    & 18.37    & 8.69    & 13.81  & 12.11 & 16.00     & 28.71  \\
 & Llama-2-13b+lora(BI) & 3.68     & 6.67   & 31.75 & 1.40   & 5.09    & 2.39     & 49.08 & 1.83   & 13.63    & 6.69     & 14.04     & 17.65   & 7.43   & 1.84   & 2.45  & 4.45    & -        & -       & -      & -     & 3.09   & 14.48 \\
 & Llama-2-13b+lora(Multi) & 45.43    & 45.25  & 38.80  & 43.00    & 36.80    & 40.06    & 49.77 & 29.89  & 30.14    & 43.16    & 34.75     & 40.99   & 16.45  & 47.50   & 34.71 & 37.11   & 33.03    & 14.11   & 31.95  & 31.89 & 33.15  & 49.70 \\
 & Llama-2-13b+FF+lora(Multi) & 19.76    & 19.98  & 18.74 & 18.95 & 18.68   & 17.04    & 24.57 & 15.70   & 16.53    & 19.70    & 17.63     & 19.36   & 15.68  & 20.79  & 17.26 & 16.71   & 17.49    & 12.69   & 18.59  & 17.60  & 16.33  & 20.8 \\
 & Mistral-7B-v0.1+lora(Multi) & 27.16    & 17.34  & 25.40  & 28.29 & 21.90    & 3.70      & 27.60  & 8.48   & 22.19    & 30.57    & 8.24      & 25.55   & 8.30    & 32.84  & 12.60  & 3.49    & 25.06    & 4.19    & 14.88  & 8.47  & 6.49   & 37.17 \\ \midrule
 
\multirow{11}{*}{\textit{\textbf{IN22\_gen}}} & 49.10     & 54.40   & 27.70  & 50.70  & 41.90    & 54.00       & 59.80  & 54.20   & 38.50     & 51.50     & 48.70      & 52.80    & 16.20   & 56.80   & 50.50  & 54.00      & 46.90     & 19.50    & 42.70   & 44.00    & 49.10   & 59.50 \\
 & IndicTrans-2 & \textbf{68.00}       & \textbf{66.40}   & \textbf{62.60}  & \textbf{74.10}  & \textbf{60.20}    & 68.40     & 66.70  & 66.30   & \textbf{62.20}     & \textbf{67.90}     & \textbf{66.10}     & \textbf{66.70}    & \textbf{61.60}   & \textbf{71.00}     & \textbf{68.60}  & \textbf{64.90}    & \textbf{57.10}     & \textbf{49.40}    & \textbf{57.70}   & \textbf{62.00}    & \textbf{66.70}   & \textbf{74.60} \\
 & Google Translate & 67.10     & 66.30   & 21.70  & 69.00    & 60.00      & \textbf{68.50}     & \textbf{67.20}  & \textbf{66.70}   & 32.70     & 66.20     & 65.10      & 66.50    & 52.70   & 70.70   & 66.40  & 64.80    & 56.00       & 00.30     & 42.80   & 61.90  & 66.40   & 73.80 \\
 & LTRC, IIIT-H & -        & -      & -     & -     & -       & -        & 54.40  & 44.30   & -        & -        & 42.70      & 47.10    & -      & -      & -     & -       & -        & -       & -      & -     & 43.90   & - \\
 & SeamlessM4T & 65.60     & 63.80   & -     & -     & -       & 66.80     & 64.90  & 64.70   & -        & 65.30     & 63.80      & 63.60    & 1.40    & 67.80   & 66.00    & 63.10    & -        & 39.60    & 38.60   & 60.00    & 64.50   & 71.10 \\
 & Llama-2-7b+lora(BI) & 4.77     & 12.80   & 23.56 & 30.80  & 22.03   & 18.59    & 35.50  & 9.40    & 24.24    & 19.09    & 15.89     & 24.88   & 5.40    & 18.53  & 6.74  & 15.03   & 21.20     & 00.46    & 24.61  & 9.55  & 7.15   & 29.47 \\
 & Llama-2-7b+lora(Multi) & 33.38    & 25.50   & 30.40  & 35.91 & 28.11   & 17.70     & 23.86 & 28.24  & 29.00       & 36.17    & 22.40      & 28.40    & 15.67  & 37.65  & 26.45 & 21.11   & 30.45    & 9.97    & 25.74  & 21.48 & 25.37  & 41.56 \\
 & Llama-2-13b+lora(BI) & 13.08    & 15.90   & 24.09 & 6.2   & 15.07   & 12.06    & 52.08 & 4.49   & 22.17    & 16.95    & 24.00        & 26.70    & 13.80  & 4.29   & 11.64 & 10.76   & -        & -       & -      & -     & 8.21   & 33.71 \\
 & Llama-2-13b+lora(Multi) & 45.95    & 45.74  & 41.47 & 49.39 & 42.73   & 36.27    & 54.26 & 38.06  & 38.01    & 49.69    & 37.81     & 49.00      & 18.81  & 53.03  & 38.26 & 33.04   & 39.50     & 14.29   & 38.00     & 36.95 & 36.30   & 53.29 \\
 & Llama-2-13b+FF+lora(Multi) & 16.70     & 17.10   & 16.83 & 17.88 & 16.76   & 14.98    & 20.50  & 15.31  & 15.95    & 17.38    & 16.60      & 17.81   & 13.40   & 17.78  & 14.66 & 15.05   & 15.84    & 12.46   & 17.68  & 16.61 & 14.87  & 17.45 \\
 & Mistral-7B-v0.1+lora(Multi) & 26.80     & 23.09  & 30.43 & 35.65 & 28.46   & 5.60      & 28.00    & 16.30   & 29.10     & 35.29    & 13.18     & 30.96   & 13.76  & 34.00     & 14.30  & 6.18    & 32.03    & 2.44    & 21.60   & 16.04 & 13.30   & 41.19 \\ \midrule
 
\multirow{11}{*}{\textit{\textbf{flores200-dev}}} & GPT-3.5 & 44.50     & 55.10   & -     & -     & -       & 53.50    & 61.50  & 52.30   & 37.40     & 52.00       & 50.50      & 51.70    & 00.00      & 55.80   & 48.90  & 56.20    & 43.00       & 19.20    & -      & 44.00    & 49.50   & 57.40 \\
 & IndicTrans-2 & \textbf{60.70}     & 65.60   & -     & -     & -       & 68.20     & 69.80  & 64.50   & \textbf{63.90}     & \textbf{72.00}       & 66.10      & 66.30    & -      & 70.40   & 67.80  & 70.40    & \textbf{54.00}       & 43.90    & -      & 64.00    & 68.90   & 65.30 \\
 & Google Translate & 60.40     & \textbf{66.00}     & -     & -     & -       & \textbf{69.00}       & \textbf{70.70}  & \textbf{64.80}   & 37.90     & \textbf{72.00}       & \textbf{66.70}      & \textbf{67.00}      & -      & \textbf{70.90}   & \textbf{67.90}  & \textbf{70.50}    & 53.60     & 00.30     & -      & \textbf{64.60}  & \textbf{69.30}   & \textbf{66.80} \\
 & LTRC, IIIT-H & -        & -      & -     & -     & -       & -        & 58.30  & 47.60   & -        & -        & 48.60      & 51.10    & -      & -      & -     & -       &          & -       & -      & -     & 50.30   & -  \\
 & SeamlessM4T & 59.80     & 64.50   & -     & -     & -       & -        & -     & 63.40   & -        & 69.90     & 65.00        & 65.10    & -      & 68.70   & 66.20  & 68.80    & -        & \textbf{44.20}    & -      & 62.80  & 67.30   & 64.90 \\
 & Llama-2-7b+lora(BI) & 3.86     & 13.27  & -     & -     & -       & 18.55    & 33.91 & 9.56   & 24.35    & 18.38    & 17.20      & 25.13   & 3.86   & 19.77  & 7.23  & 21.16   & 24.11    & 0.45    & 17.26  & 6.68  & 7.54   & 25.05 \\
 & Llama-2-7b+lora(Multi) & 32.47    & 27.27  & -     & -     & -       & 16.34    & 17.14 & 28.70   & 28.41    & 36.03    & 24.85     & 28.90    & 20.18  & 35.44  & 27.29 & 21.59   & 27.56    & 12.37   & 14.71  & 20.9  & 26.70   & 34.75 \\
 & Llama-2-13b+lora(BI) & 10.94    & 19.00     & -     & -     & -       & 10.56    & 53.87 & 5.00      & 20.58    & 17.03    & 25.83     & 28.57   & 9.73   & 4.21   & 10.39 & 8.63    & -        & -       & -      & -     & 7.70    & 21.78 \\
 & Llama-2-13b+lora(Multi) & 43.69    & 47.68  & -     & -     & -       & 37.91    & 56.27 & 38.65  & 36.59    & 51.53    & 39.66     & 47.90    & \textbf{26.15}  & 50.98  & 38.57 & 41.27   & 37.78    & 15.88   & \textbf{25.85}  & 39.00    & 39.83  & 49.76  \\
 & Llama-2-13b+FF+lora(Multi) & 18.77    & -      & -     & -     & -       & 17.44    & 21.33 & 17.45  & 17.23    & 19.26    & 17.04     & 19.97   & 18.10   & 19.00     & 16.73 & 16.43   & 17.73    & 14.26   & 16.26  & 18.31 & 17.40   & -  \\
 & Mistral-7B-v0.1+lora(Multi) & 27.96    & 25.19  & -     & -     & -       & 6.97     & 27.01 & 19.06  & 28.60     & 38.69    & 17.34     & 35.26   & 19.00     & 38.19  & 15.10  & 6.03    & 29.75    & 2.85    & 19.23  & 16.91 & 13.80   & 37.69 \\ \midrule
 
\multirow{11}{*}{\textit{\textbf{flores200-devtest}}} & GPT-3.5 & 43.90     & 54.10   & -     & -     & -       & 52.90     & 60.90  & 51.40   & 37.70     & 51.20     & 50.10      & 51.90    & 00.00      & 54.80   & 47.10  & 54.20    & 43.00       & 19.10    & -      & 43.20  & 46.90   & 55.60 \\
 & IndicTrans-2 & 59.40     & 64.80   & -     & -     & -       & 68.80     & 69.10  & 63.40   & \textbf{61.80}     & \textbf{71.30}     & 66.20      & 66.60    & -      & 69.90   & \textbf{66.80}  & 67.90    & \textbf{54.00}       & 42.30    & -      & 63.20  & 68.00     & 64.10 \\
 & Google Translate & \textbf{59.80}     & \textbf{65.30}   & -     & -     & -       & \textbf{69.70}     & \textbf{69.70}  & \textbf{63.90}   & 37.20     & 70.50     & \textbf{66.40}      & \textbf{67.20}    & -      & \textbf{70.50}   & 65.90  & \textbf{68.70}    & 53.80     & 00.30     & -      & \textbf{63.60}  & \textbf{68.50}   & \textbf{65.60}  \\
 & LTRC, IIIT-H & -        & -      & -     & -     & -       & -        & 58.40  & 47.00     & -        & -        & 48.20      & 50.90    & -      & -      & -     & -       & -        & -       & -      & -     & 49.40   & -     \\
 & SeamlessM4T & 58.30     & 63.80   & -     & -     & -       & -        & -     & 62.50   & -        & 68.50     & 65.20      & 65.00      & -      & 68.20   & 65.00    & 67.00      & -        & \textbf{42.40}    & -      & 61.40  & 66.90   & 63.90 \\
 & Llama-2-7b+lora(BI) & 3.96     & 12.63  & -     & -     & -       & 17.54    & 31.87 & 7.95   & 23.86    & 19.30     & 17.85     & 25.35   & 03.57   & 19.97  & 07.09  & 21.16   & 23.64    & 0.46    & 17.78  & 05.69  & 07.17   & 21.69 \\
 & Llama-2-7b+lora(Multi) & 30.6     & 26.28  & -     & -     & -       & 16.24    & 16.2  & 28.86  & 28.14    & 35.87    & 24.17     & 28.17   & 20.11  & 36.6   & 26.4  & 21.37   & 27.3     & 11.38   & 14     & 21.27 & 25.53  & 36 \\
 & Llama-2-13b+lora(BI) & 9.87     & 18.36  & -     & -     & -       & 9.38     & 53.3  & 4.93   & 19.71    & 15.73    & 25.7      & 28      & 8.94   & 3.88   & 9.74  & 8.59    & -        & -       & -      & -     & 7.07   & 20.7 \\
 & Llama-2-13b+lora(Multi) & 42.67    & 45.51  & -     & -     & -       & 36.97    & 55.44 & 38.15  & 35.4     & 51.23    & 39.66     & 48.08   & \textbf{25.77}  & 50.25  & 37.08 & 39.84   & 37.93    & 15.83   & \textbf{25.78}  & 37.28 & 39.18  & 48.76 \\
 & Llama-2-13b+FF+lora(Multi) & 17.96    & 19.29  & -     & -     & -       & 16.86    & 20.13 & 16.95  & 17.08    & 19.06    & 16.71     & 19.33   & 17.84  & 18.45  & 16.27 & 16.09   & 17.2     & 13.71   & 15.83  & 17.98 & 16.68  & 18.23 \\
 & Mistral-7B-v0.1+lora(Multi) & 25.7     & 23.29  & -     & -     & -       & 6.03     & 25.07 & 17.79  & 28.05    & 35.57    & 15.76     & 33.4    & 19.29  & 36.31  & 15.47 & 5.54    & 29.95    & 2.91    & 19.26  & 14.98 & 14.18  & 35.23 \\ \midrule
 
\multirow{11}{*}{\textit{\textbf{newstest2019}}} & GPT-3.5 & -        & 51.8   & -     & -     & -       & 50.7     & 56.7  & 48.1   & -        & -        & 45.1      & 48      & -      & -      & -     & 50.8    & -        & -       & -      & 40    & 43     & 19.4 \\
 & IndicTrans-2 & -        & \textbf{64.5}   & -     & -     & -       & 67.7     & \textbf{64.6}  & 62.4   & -        & -        & 61.1      & 63.9    & -      & -      & -     & \textbf{65.8}    & -        & -       & -      & 58.3  & 56.8   & 19  \\
 & Google Translate & -        & \textbf{64.5}   & -     & -     & -       & \textbf{67.8}     & -     & \textbf{62.7}   & -        & -        & \textbf{61.6}      & \textbf{64.1}    & -      & -      & -     & -       & -        & -       & -      & \textbf{58.7}  & \textbf{57.3}   & \textbf{19.6} \\
 & LTRC, IIIT-H & -        & -      & -     & -     & -       & -        & 47.9  & 44.7   & -        & -        & 44        & 48.9    & -      & -      & -     & -       & -        & -       & -      & -     & 38.8   & - \\
 & SeamlessM4T & -        & 62.3   & -     & -     & -       & 65.8     & 63.4  & 60.5   & -        & -        & 59.3      & 61.8    & -      & -      & -     & 63.9    & -        & -       & -      & 56.9  & 55.5   & 19.4  \\
 & Llama-2-7b+lora(BI)  & -        & 13.28  & -     & -     & -       & 17.13    & 27.87 & 9.74   & -        & -        & 17.53     & 25.14   & -      & -      & -     & 18.37   & -        & -       & -      & 7.01  & 8.35   & 8.24 \\
 & Llama-2-7b+lora(Multi) & -        & 26.65  & -     & -     & -       & 16.53    & 27.36 & 25.11  & -        & -        & 22.56     & 30.49   & -      & -      & -     & 20.14   & -        & -       & -      & 19.64 & 22.55  & 14.88 \\
 & Llama-2-13b+lora(BI) & -        & 25.58  & -     & -     & -       & 14.2     & 50.06 & 4.56   & -        & -        & 26.76     & 29.23   & -      & -      & -     & 8.58    & -        & -       & -      & -     & 6.08   & 10.67 \\
 & Llama-2-13b+lora(Multi)  & -        & 43.9   & -     & -     & -       & 34.31    & 52.21 & 34.5   & -        & -        & 34.56     & 44.86   & -      & -      & -     & 34.35   & -        & -       & -      & 33.24 & 32.65  & 17.51\\
 & Llama-2-13b+FF+lora(Multi) & -        & 17.11  & -     & -     & -       & 15.23    & 18.4  & 15.5   & -        & -        & 15.49     & 16.98   & -      & -      & -     & 15.05   & -        & -       & -      & 16.48 & 15     & 13.28 \\
 & Mistral-7B-v0.1+lora(Multi) & -        & 24.9   & -     & -     & -       & 6.06     & 26.04 & 18.47  & -        & -        & 14.09     & 30.3    & -      & -      & -     & 4.6     & -        & -       & -      & 13.91 & 13.1   & 14.03 \\ \bottomrule
\end{tabular}%
}
\caption{chrF scores across Models and Benchmark-sets; 22 Indian Languages to English;\\The symbol `-' indicates that the benchmark dataset for a particular language or machine translation system was not available during the evaluation period.  Here, LORA stands for Low-Rank Adaptation of Large Language Models based fine-tuning. Multi stands for the multilingual model, FF for full-finetuning, and FF+lora stands for 2-stage fine-tuning.}
\label{tab:xx_en_chrf}
\end{table*}
\end{document}